\documentclass{article} %

\usepackage[final]{colm2026_conference}

\usepackage{microtype}
\usepackage{hyperref}
\usepackage{url}
\usepackage{booktabs}

\usepackage{amsmath}
\usepackage{amssymb}
\usepackage{mathtools}
\usepackage{amsthm}

\usepackage[capitalize,noabbrev]{cleveref}

\theoremstyle{plain}

\theoremstyle{definition}

\theoremstyle{remark}

\usepackage[textsize=tiny]{todonotes}

\usepackage{multirow}
\usepackage{xspace}
\usepackage{wrapfig}
\usepackage{enumitem}
\usepackage{makecell}
\usepackage{caption}
\usepackage{subcaption}
\usepackage[table]{xcolor}

\usepackage[most]{tcolorbox}
\usepackage{etoolbox}
\tcbuselibrary{listingsutf8}

\newtcolorbox{LLMPrompt}[1][]{
  enhanced, breakable, verbatim,
  colback=gray!5, colframe=black!15, boxrule=0.4pt, arc=1.5mm,
  left=6pt,right=6pt,top=6pt,bottom=6pt,
  fontupper=\small\ttfamily,
  colbacktitle=gray!50,
  title=#1
}

\iftrue  %
    \newcommand{\draftcomment}[3]{}
\else
    \newcommand{\draftcomment}[3]{\textcolor{#2}{{\bf\small [#1: #3]}}}
    
\fi

\newcommand{\eg}{\hbox{\emph{e.g.,}}\xspace}
\newcommand{\ie}{\hbox{\emph{i.e.,}}\xspace}

\newcommand{\cost}{\textit{Cost}}
\newcommand{\perf}{\textit{Perf}}
\newcommand{\code}[1]{{\tt {\small #1}}}

\newcommand{\method}{\textsc{HandRaiser}\xspace}
\newcommand{\speaker}{speaker\xspace}
\newcommand{\listener}{listener\xspace}
\newcommand{\speakers}{speakers\xspace}
\newcommand{\listeners}{listeners\xspace}

\definecolor{mediumgreen}{RGB}{0, 150, 50}

\usepackage{lineno}

\definecolor{darkblue}{rgb}{0, 0, 0.5}
\hypersetup{colorlinks=true, citecolor=darkblue, linkcolor=darkblue, urlcolor=darkblue}

\title{Learning to Interrupt in Language-based Multi-agent \\ Communication}

\author{%
  Danqing Wang$^1$\thanks{Majority of the work done during an internship at Meta FAIR.} 
  \quad Da Yin$^2$
  \quad Ruta Desai$^2$
  \quad Lei Li$^1$ 
  \quad Asli Celikyilmaz$^2$
  \quad Ansong Ni$^2$ \\
  $^1$CMU, $^2$Meta FAIR\\
  \texttt{danqingw@andrew.cmu.edu, 
  ansongni@meta.com}
}

\begin{document}

\ifcolmsubmission
\linenumbers
\fi

\maketitle
\setlength{\dbltextfloatsep}{8pt plus 2pt minus 2pt}
\setlength{\dblfloatsep}{8pt plus 2pt minus 2pt}

\begin{abstract}
When a colleague starts explaining something you already understand, you interrupt them. This simple act, a listener
taking control of the conversation, is natural in human communication but absent in current verbose LLM multi-agent systems. Current approaches address communication efficiency only from the speaker side, compressing messages before they are sent. We flip the perspective: rather than making speakers
more concise, we let listeners decide when they have heard enough.
We propose a new communication paradigm in which the \listener can interrupt the \speaker mid-generation. We find that LLMs, given this ability, are overconfident and interrupt too early before receiving sufficient information.
This finding motivates \method, a learning method that predicts the right moment to interrupt based on estimated future reward and communication cost.
We evaluate our framework on three multi-agent tasks: 2-agent text pictionary, 3-agent meeting scheduling, and 3-agent
debate. \method reduces communication cost by 32.2\% over the non-interruptible baseline while achieving comparable or
superior task performance, with interruption behavior that generalizes across different agents and tasks.

\end{abstract}

\section{Introduction}
Large language model multi-agent systems have shown remarkable performance in \textit{reasoning}~\citep{du2023improving,zhuge2024language,qian2025scaling}, \textit{software engineering}~\citep{hu2025selfevolving,he2025llm}. They also show potential in simulated social environments such as \textit{strategic games}~\citep{xu2023exploring,Wang2023AvalonsGO} and \textit{social behavior modeling}~\citep{park2023generative, dubois2023alpacafarm}.
Compared with single-agent systems that rely on one LLM to complete a task sequentially, multi-agent systems can complete more complex, long-horizon tasks at scale, as demonstrated by recent deployments such as OpenClaw's personal assistant~\citep{openclaw2026}, Claude Code's agentic research workflows~\citep{anthropic2025agent}, and Kimi Agent Swarm~\citep{team2026kimi}.

However, multi-agent systems suffer from a fast-growing context due to the verbosity of typical LLM-generated messages and the number of messages being sent across different agents.
Such inefficient communication not only decreases the general performance due to overloaded context~\citep{guo2024large,cemri2025multi}, but also increases the amount of compute and incurs larger latency at inference time~\citep{wang-etal-2025-agentdropout,zhang2025cut}.

\begin{figure*}[t]
    \centering
    \begin{subfigure}[h]{0.55\linewidth}
        \centering
        \includegraphics[width=1\linewidth]{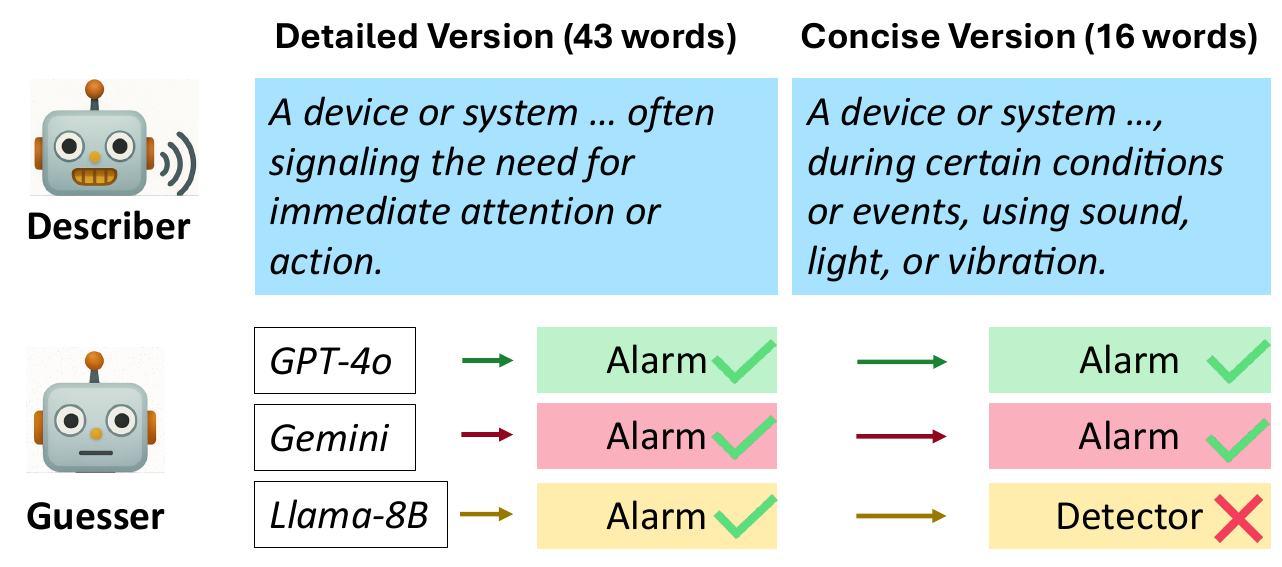}
        \caption{The speaker-oriented compression can not fit listeners with different understanding.}
        \label{fig:speaker}
    \end{subfigure}
    \hfill
    \begin{subfigure}[h]{0.42\linewidth}
        \centering
        \includegraphics[width=1\linewidth]{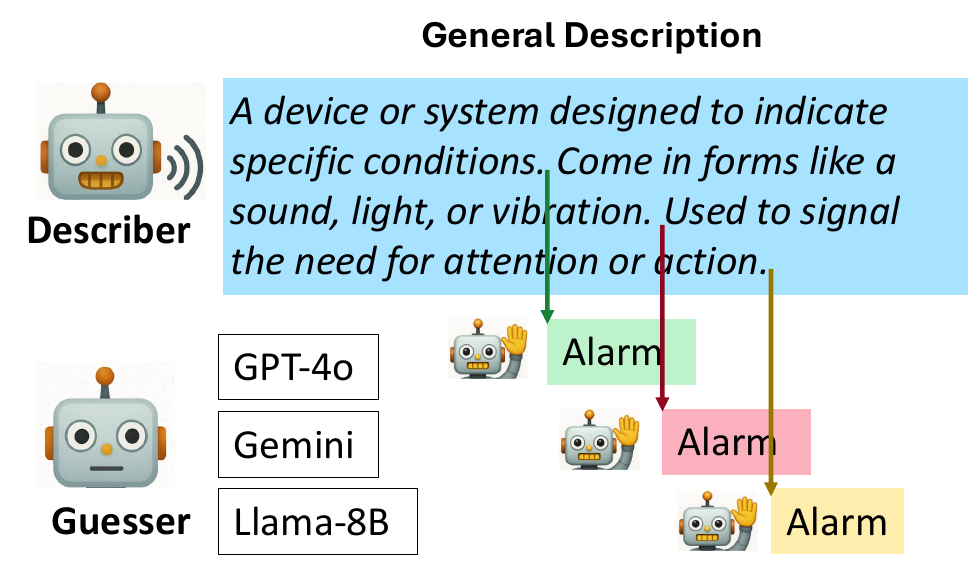}
        \caption{We let the listeners themselves to decide when to interrupt and respond.}
        \label{fig:listener}
    \end{subfigure}
    \caption{Text Pictionary: the speaker describes words for the listener to guess.}
    \label{fig:main}
\end{figure*}

Consider a single interaction (\ie message) in a multi-agent system. It involves two types of roles: the \speaker who sends the message, and the \listener who receives it; and efficient communication should aim to convey the \speaker's intent to the \listener with minimal words.
Towards efficient multi-agent communication, previous work mostly focuses on prompting or training the \speaker to generate more succinct messages. This can be seen as \speaker-oriented compression~\citep{fang2025thinkless,qiao2025concise}.
While such speaker-oriented compression is initially helpful in reducing overall verbosity, further compression becomes significantly challenging. It requires the speaker to identify the parts that are indispensable to the listener. More importantly, such parts may also vary for different listeners.
Take the ``Text Pictionary'' game shown in \autoref{fig:main} as an example. In this game, one agent describes a secret entity and the other agent guesses the answer. As shown on the left side (\autoref{fig:speaker}), given the same 16-word description, some agents can already guess the word while other agents are still struggling.

Rather than compressing from the \speaker side, we flip the perspective: we let \listeners decide when they have heard enough. Such interruption naturally occurs in human conversation, where people interrupt to skip redundant information or signal readiness to respond~\citep{lycan1977conversation,bennett1978interruptions,ng1995interruption}.
In this work, we introduce an \textit{interruptible communication framework} for multi-agent LLM systems. For each interaction, the \speaker sends tokens to the \listener in chunks. The \listener decides whether to interrupt, skipping information it deems redundant or asking for clarification when confused. Upon interruption, the \speaker halts and awaits the \listener's response.
In contrast to speaker-oriented communication in \autoref{fig:speaker}, the interruption mechanism in \autoref{fig:listener} allows different \listeners (\ie, guessers in this game) to interrupt at different points without requiring the \speaker to tailor its messages.

To teach LLMs when to interrupt, we investigate both prompting-based and learning-based methods. We find that LLMs given this ability are overconfident, interrupting too early before receiving sufficient information. This overconfidence motivates \method, a learning method that samples multi-agent rollouts, annotates each potential interruption point based on its \textit{payoff} (estimated future token reduction without performance loss), and finetunes the agent to predict interruption points with higher task reward and lower communication cost.

We evaluate our interruptible communication framework and trained \method model with multiple LLMs (\code{Llama-3.1}, \code{GPT-4o}, and \code{Gemini-2.0-flash}) across three multi-agent scenarios (Text Pictionary, Meeting Scheduling, and MMLU-Pro Debate). These tasks are selected based on three principles: (i) communication-dependence, where the task relies on inter-agent communication; (ii) necessity of multi-agent setup, such as multi-player games and agents with private information; and (iii) coverage of common multi-agent task types, such as debate for reasoning.
The three tasks collectively span 2-agent and 3-agent configurations, game-style and real-world scenarios, fixed speaking order and moderator-driven topologies, and tasks with private information that cannot be globally shared.

Experimental results show that \method reduces communication cost by 32.2\% over the non-interruptible baseline while achieving comparable or superior task performance. Further analysis shows that the learned interruption behavior transfers to other speaker types and tasks, and generalizes to more complex multi-agent communication patterns.

\section{Interruptible Communication Mechanism}

In this section, we introduce an interruptible communication framework that operates through a \listener-oriented decision process, which dynamically evaluates whether to stop the ongoing generation of the \speaker and start responding. 

\subsection{Interruptible Communication Protocol}
\label{sec:protocol}

\begin{figure*}
    \centering
    \includegraphics[width=0.95\linewidth]{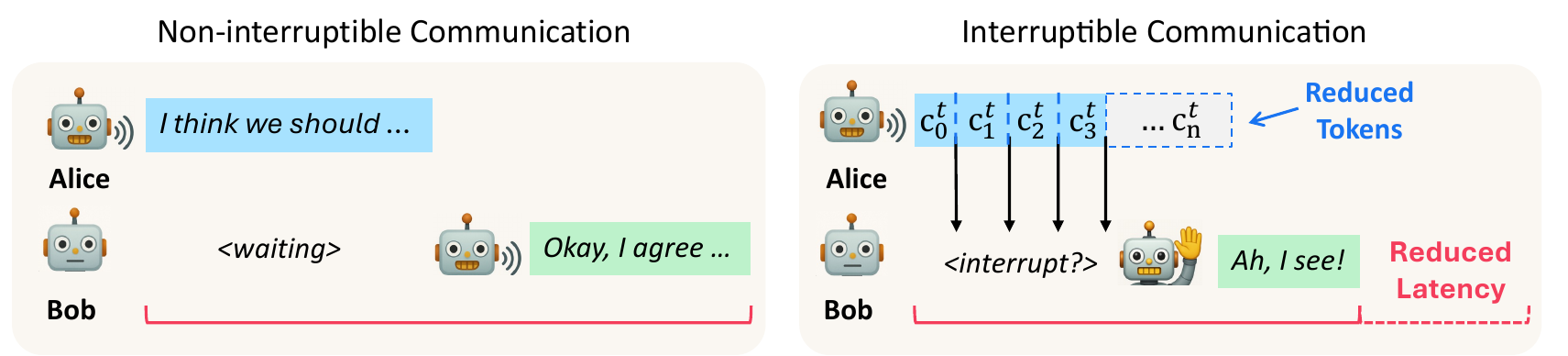}
    \caption{In non-interruptible communication (left), one agent (Bob) must wait for the completion of a full generation from another agent (Alice) to start responding. However, in the interruptible case, the \speaker Alice sends messages to the \listener Bob in chunks, and Bob will decide whether to interrupt. Upon interruption, Alice halts, and Bob responds.}
    \label{fig:overview}
\end{figure*}

\textbf{Basic formulation.} Consider one back-and-forth round of interaction illustrated in \autoref{fig:overview}, with Alice $\mathcal{A}$ speaking first and Bob $\mathcal{B}$ responding afterwards.
With the current multi-agent communication framework (illustrated on the left), Bob waits for Alice to finish speaking, and only after Alice has generated and sent the full message can Bob begin its response.
In our proposed interruptible communication protocol (illustrated on the right of \autoref{fig:overview}),
the \speaker Alice generates its response $X_t$ in fixed-sized chunks\footnote{We use the notion of chunks to strike a balance between the granularity of interruption points and amount of interruption predictions needed. And we treat their sizes (a single token $\sim$ full message) as a hyperparameter.},
\ie $X^t=\{c_0^t, \cdots, c_n^t\}$, and each chunk $c_i^t$ is transmitted to the \listener after they are generated.
Upon receiving each chunk $c_i^t$, the \listener Bob decides whether it wants to interrupt based on the previous chat history, as well as the prefix of the message by accumulating all the chunks it received so far in this round, \ie $\hat{X}^t=\{c_0^t, \cdots c_i^t\}$. 
When Bob decides to interrupt, it sends an interruption signal to Alice and begins generating its response.
Upon receiving this signal, Alice halts its generation and awaits Bob's response.

\textbf{Communication patterns in multi-agent systems} Multi-agent communication processes can be decomposed into multiple such basic communication processes. Generally, communication in $k$-agent systems can be categorized into (i) \textit{A fixed speaking order conversation} (\eg round-robin): the agents speak in a predetermined order;  (ii) \textit{A free discussion} (\eg, group chat): the current agent broadcasts to all other agents. 
In the first scenario, given the speaking order $\{a_1, a_2, \cdots, a_{k}\}$, the communication process can be decomposed into $k$ independent basic communication processes between $a_i \rightarrow a_{i+1}$, where $a_i$ is the speaker and $a_{i+1}$ is the listener. $a_{i+1}$ can interrupt $a_{i}$ during its speaking. 
In the second scenario, where the current speaker $a_{i}$ broadcasts to the rest, other agents decide independently whether to interrupt. This decomposes message broadcasting to multiple parallel basic communication processes. 
If two agents might both want to interrupt, we refer to common resolutions of this, such as first-come, first-served. The decomposition makes it possible to handle the conflicting or cascading interruptions. 
More detailed explanations are in Appendix \ref{sec:app-discussion}.

This interruptible mechanism can improve the communication efficiency in two ways: (i) The generation cost of Alice can be saved once its generation is halted; and (ii) the communication latency is reduced because Bob does not need to wait the full message and can potentially respond earlier. 

\subsection{Evaluating the Payoff of Interruptions}
\label{sec:payoff}

Akin to the compression-fidelity tradeoff in information theory~\citep{shannon1959coding}, an interruption in communication also constitutes a tradeoff between communication cost and quality, for which we measure as total number of generated tokens and final task performance, respectively.
\paragraph{Estimating communication cost and quality.} 
Given a set of $m$ agents where $A = \{a_1, \cdots, a_m\}$, a conversation $C_A^{1:T} = \{X_{a_1}^1, X_{a_2}^1, \cdots, X_{a_m}^1, \cdots, X_{a_m}^T\}$ denotes $T$ rounds of messages where each message $X_a^t$ is the message generated by agent $a$ at round $t$.
Following previous work in optimizing multi-agent communication~\citep{wang-etal-2025-agentdropout,zhang2025cut}, we estimate the communication cost by
counting the total amount of tokens generated by all agents till the end of the conversation, \ie $\textit{Cost}(C_A^{1:T})=\sum_{a\in A}\sum_{t=1}^{T} |X_a^t|$.
And to measure the overall communication quality, we use goal-oriented task performance of respective tasks, \ie $\textit{Perf}(C_A^{1:T})\in [0,1]$, to measure the communication quality.

\paragraph{Balancing the cost-quality tradeoff.}
When considering whether it is ideal to interrupt after receiving a certain chunk $c_i^t$ for agent $a$, it is important to note that it not only affects the current turn $X_a^t$ at round $t$, but also every turn afterwards. This is because the truncated message ${\hat{X}}_a^t$ from agent $a$ will be factorized as the context for all subsequent interactions ${\hat{C}}_A^{t:\hat{T}}$. 
It may even end at a different round $\hat{T}$ depending on the ending criteria. 
Thus, for the interruption of agent $a$ at $t$-th round after its $i$-th chunk, we formulate the potential changes (\ie $\Delta$), compared with no interruption, in cost of communication and task performance as:\footnote{All the expectations $\mathbb{E}(\cdot)$ here are taken over all possible conversations stemming from the same prefix.}
\begin{equation*}
\begin{aligned}
\Delta_{\mathrm{cost}}(a,t,i)
&= -\sum_{k=i+1}^n |c_k^t|
 + \mathbb{E}[\mathrm{cost}(\hat{C}_A^{t:\hat{T}})]
 - \mathbb{E}[\mathrm{cost}(C_A^{t:T})] \\
\Delta_{\mathrm{perf}}(a,t,i)
&= \mathbb{E}[\mathrm{perf}(C_A^{1:{t-1}} \parallel \hat{X}_a^t \parallel \hat{C}_A^{t:\hat{T}})]
 - \mathbb{E}[\mathrm{perf}(C_A^{1:T})]
\end{aligned}
\end{equation*}
where $\parallel$ denotes the concatenation of turns.
An ideal interruption would aim to reduce communication cost, \ie $\min \Delta_{\cost}(a, t, i)$, while maximizing the task performance, \ie $\max \Delta_{\perf}(a, t, i)$. 
And to combine these two objectives, we consider an interruption to have \textit{positive payoff} when it reduces the communication cost while maintaining the task performance, \ie $\Delta_{\cost}(a, t, i) < 0 \;\land\; \Delta_{\perf}(a, t, i) \geq 0$.

\section{Learn to Interrupt with Language Models}

\begin{figure*}[t]
    \centering
    \includegraphics[width=1\linewidth]{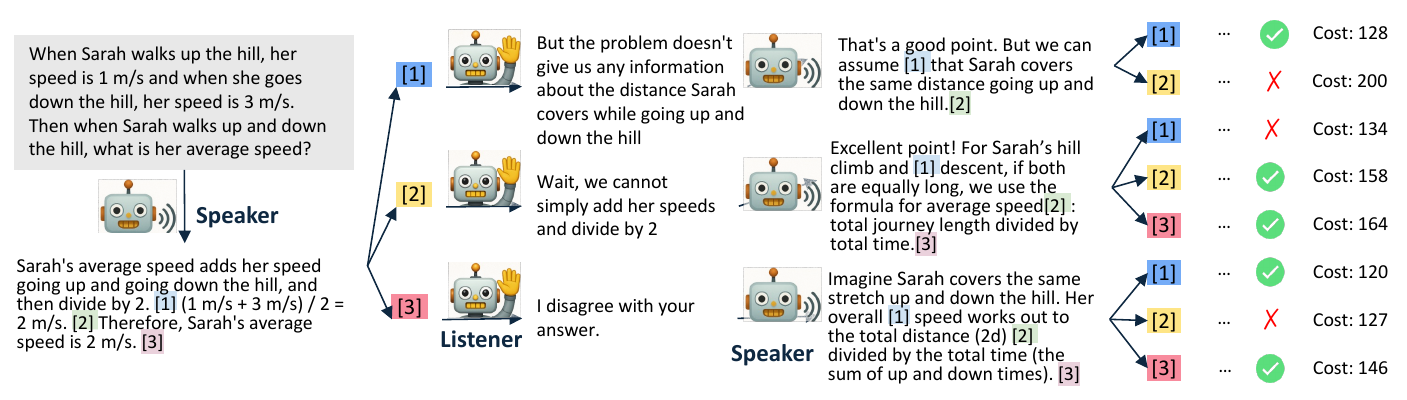}
    \caption{Sampling to estimate the expectation of communication cost and task performance. 
    }
    \label{fig:sampling}
\end{figure*}

\subsection{Prompt-based Method}
\label{sec:prompt}
An intuitive way to model the interruption is to let the agent decide whether to interrupt by prompting. Specifically, when Alice speaks at the round $t$, for each segment $c_i^t$, Bob generates an interruption response (Yes or No) to indicate whether he will interrupt based on the chat history $C_{A}^{1:t-1}$ and the chunks $c_{1:i}^t$ he received so far in the current round. If the answer is yes, Bob will send an interruption signal to Alice and generate its response immediately afterwards. If the answer is no, Bob will wait for the next chunk and repeat this process until Alice completes its message. Alice will stop its generation when it receives the interrupt signal from Bob. A simple prompt template is shown in \autoref{sec:app-evaluation}. We refer to this as Prompt-based interruption.

However, we find that by default the agent struggles to find a suitable interruption point, as it tends to be overconfident and interrupts before receiving enough information. This can reduce the cost of the current round but often leads to more rounds overall or lower task performance. We discuss this in \autoref{sec:result}.

\subsection{Learning Good Interruption Points}
\label{sec:learn}

\textbf{Estimating with tree sampling.}
The agent needs to know whether the current interruption can lead to a \textit{positive payoff} compared to the full message. As described in \autoref{sec:payoff}, we need to estimate the expected communication cost  $\mathbb{E}[\cost(\hat{C}_A^{t:\hat{T}})]$ from the interruption point to the end of the communication and the task performance based on this trajectory $\mathbb{E}[\perf([C_A^{1:{t-1}} \parallel \hat{X}_a^t \parallel {\hat{C}}_A^{t:\hat{T}}])]$. Since the interruption will affect the subsequent interactions, we use tree-based sampling to simulate the future interaction, as shown in \autoref{fig:sampling}.

At round $t$, the current speaking agent (\ie, Alice) $a$'s message $X_a^t$ is segmented into chunks $\{c_1^{t}, \cdots, c_n^{t}\}$, and after each chunk $c_i^t$ is a potential interruption point. It constructs $n$ choices for interruption for this turn, and each point $i \in \{1,\dots,n\}$ is modeled as one branch for the potential interruption, resulting in different partial messages upon which the response of the listening agent (\ie Bob) is based to generate a response. Here, $i=n$ indicates no interrupt for this turn. 
We then let Bob generate a response based on the partial message. 
In the next round $t+1$, Alice will generate a new message $X_a^{t+1}$ based on Bob's response within the branch, and so on. This exploration will continue until the end of the communication, which is decided by the maximum round number or the task completion signal. For each node with the partial message from Alice, we roll out the rest of the conversation ${\hat{C}}_A^{t:\hat{T}}$  based on a random interruption policy and calculate their task performance and communication cost. We take their average as the estimated task performance and communication cost for this node.

\textbf{Labeling training data.}
We calculate $\Delta_{\cost}(a, t, i)$ and $\Delta_{\perf}(a, t, i)$ for each tree branch $i$ and label points with positive delta for both cost and performance as positive interruptions, and the rest as negative. The tree trajectory is then formatted for instruction fine-tuning using the same prompt template as the Prompt-based method, with labels 1 and 0 converted to Yes and No. Specifically, the path from the root to the node is the chat history, and the partial message in this node is the message chunks received in this round. 

We use supervised finetuning to train \method on these instruction data with cross-entropy loss to learn the interruption decision based on the payoff estimation. During inference, when the other agent is speaking in a streaming manner, \method generates a one-token interruption decision for each chunk to decide whether to interrupt.

\subsection{Inference Complexity and Cost}
\label{sec:app-inference-cost}
In this paper, following previous work in optimizing multi-agent communication~\citep{wang-etal-2025-agentdropout,zhang2025cut}, we define communication cost as communication latency in the multi-agent system, and use reduced generation tokens as a proxy for reduced latency, because:

\textbf{Extra computation of encoding and network latency can be ignored compared with the inference cost.} First, the KV cache can be used to avoid repeated computations in encoding previous chunks. The decoding phase is significantly more computationally expensive than the prefill phase, especially at batch size 1, because decoding is memory-bound: while the computation cost of an input token and an output token is theoretically similar, for each generated token, model weights must be reloaded from GPU global memory (HBM) to on-chip memory (L1 / shared memory, ~200KB per SM), which is costly. Popular API providers with advanced batching strategies and parallel architectures, such as OpenAI, still set the price of output tokens 4-10 times higher than the input tokens, indicating a significant computation gap between the input and output. Besides, LLM client and server are usually hosted on the same local network under the benchmark scenarios. The network latency is negligible (often $<1$ms). 

\textbf{Addition inference cost in \method for the interruption decision is linearly related to the number of chunks.} A smaller chunk size leads to more frequent interruption requests, with one token per request. 
Suppose the chunk size is $l_c$ and the length of the full message is $L$. If no interruption is made, then the interruption mechanism will increase $L / l_c$ additional cost for the interruption cost. If an interruption happens (at least one chunk is reduced), then reduced tokens can make up for the interruption token $L / l_c < l_c \to l_c < \sqrt{L}$. Therefore, we can select a chunk size larger than $\sqrt{L}$ to ensure the interruption can benefit the communication cost in the worst case.

\section{Experiment}
\subsection{Task Setup}
\label{sec:task-setup}
We evaluate three multi-agent environments from simulated and real-world scenarios. 
We briefly introduce these scenarios here, and more details can be found in \autoref{sec:app-task-setup}.

\begin{table*}[!htp]\footnotesize\setlength{\tabcolsep}{4pt}
\centering
\caption{Different communication methods. The ``Generic'' and ``Concise'' baselines are non-interruptible, and the others are of different interruption strategies. Results are averaged over three speaker agents, and the full results can be found in \autoref{tab:full}.
``SR'' $=$ Success Rate. 
}
\label{tab:result}
\begin{tabular}{lcccccccc}
\toprule
 & &\multicolumn{2}{c}{Text Pictionary} &\multicolumn{2}{c}{Meeting Scheduling} &\multicolumn{2}{c}{MMLU-Pro-Debate} \\
\cmidrule{3-8}
Listener & Methods &SR &Cost &SR &Cost &SR &Cost \\
\midrule
    \multirow{5}[0]{*}{\code{Llama-8B}} & Generic & $0.743_{\pm 0.033}$ & $346_{\pm 37}$ & $0.273_{\pm 0.077}$ & $1361_{\pm 86}$ & $0.520_{\pm 0.050}$ & $1531_{\pm 126}$ \\
          & Concise & $0.753_{\pm 0.023}$ & $277_{\pm 24}$ & $0.257_{\pm 0.053}$ & $1155_{\pm 80}$ & $0.523_{\pm 0.090}$ & $829_{\pm 176}$ \\
          \cmidrule{2-8}
          & Random & $0.710_{\pm 0.033}$ & $360_{\pm 22}$ & $0.197_{\pm 0.040}$ & $1760_{\pm 114}$ & $0.537_{\pm 0.050}$ & $1531_{\pm 85}$ \\
          & Prompting   & $0.500_{\pm 0.033}$ & $461_{\pm 25}$ & $0.200_{\pm 0.020}$ & $1778_{\pm 129}$ & $0.537_{\pm 0.063}$ & $1690_{\pm 130}$ \\
          & \method & $0.743_{\pm 0.023}$ & $262_{\pm 27}$ & $0.307_{\pm 0.049}$ & $1042_{\pm 83}$ & $0.583_{\pm 0.080}$ & $782_{\pm 181}$ \\
    \midrule
    \midrule
    \multirow{5}[0]{*}{\code{Llama-70B}} & Generic & $0.773_{\pm 0.023}$ & $401_{\pm 46}$ & $0.420_{\pm 0.040}$ & $1228_{\pm 52}$ & $0.647_{\pm 0.013}$ & $1617_{\pm 78}$ \\
          & Concise & $0.800_{\pm 0.027}$ & $326_{\pm 37}$ & $0.437_{\pm 0.043}$ & $1025_{\pm 82}$ & $0.657_{\pm 0.057}$ & $847_{\pm 113}$ \\
          \cmidrule{2-8}
          & Random & $0.787_{\pm 0.033}$ & $419_{\pm 31}$ & $0.420_{\pm 0.080}$ & $1228_{\pm 101}$ & $0.623_{\pm 0.043}$ & $1617_{\pm 152}$ \\
          & Prompting   & $0.663_{\pm 0.040}$ & $523_{\pm 43}$ & $0.420_{\pm 0.060}$ & $1585_{\pm 136}$ & $0.653_{\pm 0.073}$ & $1707_{\pm 190}$ \\
          & \method & $0.790_{\pm 0.020}$ & $294_{\pm 49}$ & $0.447_{\pm 0.063}$ & $1010_{\pm 87}$ & $0.657_{\pm 0.070}$ & $806_{\pm 119}$ \\
\bottomrule
\end{tabular}
\end{table*}

\textbf{Text Pictionary.}
Text Pictionary is a text version of Pictionary that consists of two agents: a \textit{describer} that describes a secret entity (\eg ``Alarm'' as in \autoref{fig:main}) without explicitly revealing it, and a \textit{guesser} trying to guess that entity.
The winning condition for both agents is for the guesser to guess the answer within a limited number of rounds.
For the interruptible communication, 
the guesser can interrupt the describer once it feels confident enough to make a guess. 
We collect 100 of these entities for evaluation.

\textbf{Meeting Scheduling.}
Scheduling meetings is a practical use case for LLM-based agents, and we follow
Natural Plan~\citep{zheng2024natural} to synthesize a challenging meeting scheduling scenario for multi-agent communication. 
It aims to schedule meetings between three agents (\ie one \textit{traveler agent} and two \textit{planning agents}) under tight meeting constraints, such as limited availability and travel time between meeting locations.  
For interruptible communication, the traveler agent can interrupt the planning agent once it has enough information to schedule meetings. 
The meeting scheduling is successful only if it satisfies all the hard constraints. 
We synthesize 50 such meeting scheduling tasks with different types of constraints.

\textbf{MMLU-Pro Debate.}
We further investigate multi-agent communication efficiency for debating on reasoning tasks.
We follow the debate setting~\citep{liang2023encouraging,khan2024debating} to set up a three-agent framework: given a reasoning problem, a \textit{positive (pro) agent} argues for a correct solution and a \textit{negative (con) agent} argues for an incorrect solution.
And the moderator asks clarifying questions in each turn, and at the end of the debate, it decides which answer is correct. 
For interruptions, the moderator can interrupt the debaters when it is confident in selecting the correct answer, and gets a binary reward for the correctness of the selected side.
For the reasoning task itself, we use the popular reasoning benchmark MMLU-Pro~\citep{wang2024mmlu}, and randomly choose a subset of 100 instances as seed questions for the debate. We also use the \code{Llama-3.1-405B-Instruct} model to generate the correct and incorrect solutions as the initial debate position. 

It is important to note that when setting up the main experiments, we only consider that one of the agents in each task can interrupt others, \ie the \textit{guesser} in text pictionary; the \textit{traveler} in meeting scheduling; and the \textit{moderator} for multi-agent debate. 
We investigate more complex communication with multiple interruptible agents in \autoref{sec:complex-interruption}. We refer to the agents that are receiving messages and can potentially interrupt others as ``\textbf{listeners}'', and the other agents that are sending messages as ``\textbf{speakers}''.

\subsection{Baselines and Implementation Details}
\textbf{Language Models.} We experiment with open-source LLMs from the \code{Llama-3.1} family (\ie \code{Llama-3.1-\{8B, 70B, 405B\}-Instruct})~\citep{grattafiori2024llama} and closed-source LLMs, \code{GPT-4o}~\citep{hurst2024gpt} and \code{Gemini-2.0-flash}~\citep{team2023gemini} as the backend. 

\textbf{Baselines.} We set up several non-interruptible and interruptible baselines for multi-agent communication. The non-interruptible baselines include a \textbf{Generic} version and a \textbf{Concise} version, the latter being a \speaker-oriented compression method that prompts the model to be more concise as shown in \autoref{fig:speaker}.
Interruptible baselines include a \textbf{Random} interruption (\ie randomly pick one point to interrupt in each turn), a \textbf{Prompt}-based interruption.

\textbf{Implementation details.}
For each setting, we evaluate 3 times using different random seeds and report the average results. 
The maximum rounds of communications are set to 10 for all three tasks, as we find in most cases the task can be completed with this budget. 
We set the default chunk size to 16 unless otherwise specified. When calculating the communication cost, we sum up generation tokens from all agents in the system, including the interruption response (one token per chunk) for interruptible baselines. 
For training, we supervise finetune the \code{LLama-3.1-\{8B,70B\}-Instruct} models with a learning rate of $1e-7$ for $500$ steps. In the main experiment, we fine-tune the interruption behavior for each task. 
The evaluation details and prompts are in \autoref{sec:app-training}.

\subsection{Main Results}
\label{sec:result}

In \autoref{tab:result}, we report the success rate and communication cost with \code{Llama-\{8B,70B\}} \listeners across three tasks. We take \code{Llama-70B}, \code{Llama-405B}, and \code{Gemini-2.0-Flash} as the \speakers and average over both metrics for more robust evaluation.

\textbf{\method achieves comparable or higher success rates with lower communication cost for both 8B and 70B listener agents}. This indicates that \method can effectively reduce communication cost without loss of task performance. Specifically, compared to the generic baseline, \method reduces cost by 24.3\% on Text Pictionary, 23.4\% on Meeting Scheduling, and 48.9\% on MMLU-Pro-Debate for the \code{Llama-8B} listener. The reduction rate for the \code{Llama-70B} listener agent is higher in  Text Pictionary and MMLU-Pro-Debate. We find that in these two scenarios, the 70B listener agent is more verbose than the 8B one. However, the 70B Llama traveler in Meeting Scheduling is more efficient than the 8B Llama in sharing information and scheduling, reducing the conversation from 14.8 to 11.5 messages in the generic baseline.

\textbf{Random and prompt-based interruption both hurt performance.} They lower success rates and raise communication costs by triggering more message rounds.
The prompt-based baseline is overconfident. For example, it interrupts in the first chunk 87.6\% of the time in meeting scheduling, even though a typical planning agent's message spans about 3.8 chunks. Cutting off early means the listener misses critical information, inflating the message count from 11.5 to 15.0. Any cost saved from one fewer message is outweighed by the extra back-and-forth. We provide a detailed discussion on the Prompting performance in \autoref{sec:analysis}.
Random interruption is more balanced in timing (13.2 messages), making it modestly better than prompting, but still worse than \method, which reaches 12.3 messages by weighing interruption payoff before acting.

\textbf{\method is shown to consistently reduce cost with comparable performance across different speaker agents.}
We also evaluate how the speaker agent affects the interruption behavior of the listener agent. We take \code{Llama-70B}, \code{Llama-405B}, and \code{Gemini-2.0-flash} as the describer and \code{Llama-8B} as the guesser and compare their performance in \autoref{fig:8b_sr} and \autoref{fig:8b_nt}. \method's learned interruption behavior can fit different types of behavior of the speaker agents. For example, Gemini is more concise and precise than the Llama models when describing the entity, and \method can still work well. The interruption behaviors of other baselines also show similar trends across different speaker agents.

\begin{table}[t]
 \begin{minipage}[h]{0.48\linewidth}
 \includegraphics[width=1\linewidth]{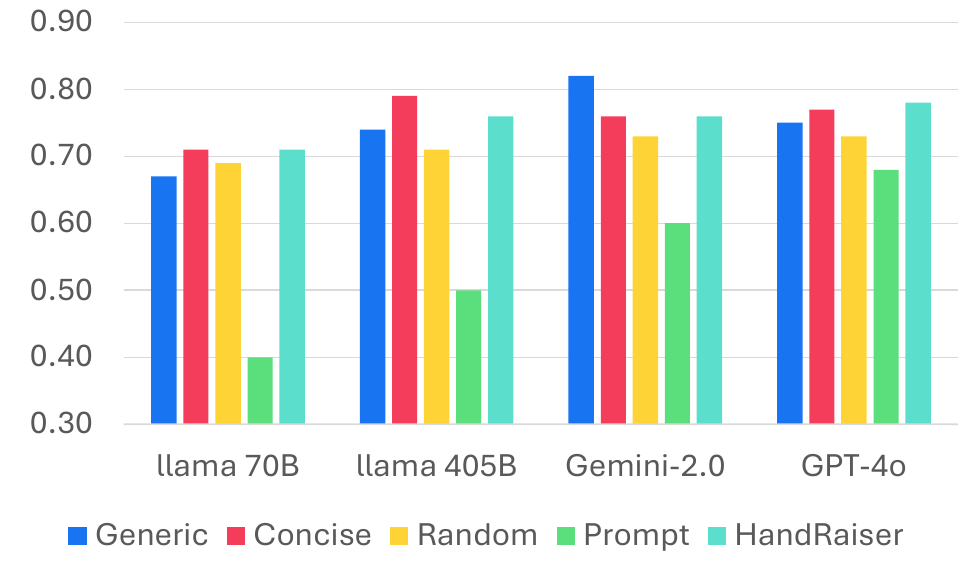}
    \captionof{figure}{Avg. Success Ratio of \code{Llama-8B} listening agent with different speaking agents.}
    \label{fig:8b_sr}
    \end{minipage}
    \hfill
    \begin{minipage}[h]{0.48\linewidth}
    \includegraphics[width=1\linewidth]{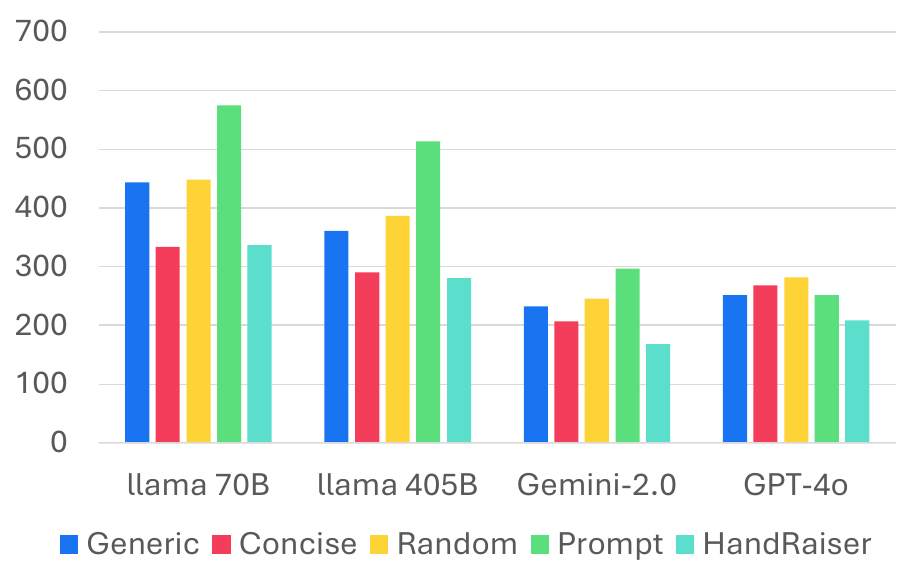}
    \captionof{figure}{Avg. comm. cost of \code{Llama-8B} listening agent with different speaking agents.}
    \label{fig:8b_nt}
    \end{minipage}
\end{table}

\subsection{What affects the interruption behaviors?}
\label{sec:analysis}
In this section, we take text pictionary as the main scenario to understand the factors that impact interruption behaviors.

\textbf{Learned interruption behavior by \method can be applied to other \speakers}
After fine-tuning \listeners on conversations with \code{Llama-70B}, \code{Llama-405B}, and Gemini, we evaluate on a new \speaker GPT-4o. Results are shown in \autoref{fig:8b_sr} and \autoref{fig:8b_nt}. \method achieves slightly better performance at a lower cost, indicating that the learned interruption behavior transfers to unseen speaker agents without additional fine-tuning.

\textbf{Why does the prompt-based interruption perform poorly?}
We investigate this by defining five understanding levels (not at all, minimal, partial, good, fully) and asking a GPT-4o \listener to self-report its understanding after each chunk. Details are described in Appendix \ref{sec:app-understanding}. As shown in \autoref{fig:un_gpt_incorrect}, even when the agent's guess is incorrect, indicating insufficient comprehension, it still rates its understanding as \textit{good} or \textit{fully} in most cases. This overconfidence leads to premature interruption: the agent believes it has grasped the describer's intent before it actually has, causing it to cut off the message too early.

\textbf{Does the interruptible communication paradigm help listener agents better understand different speakers?}
To directly address the motivation in \autoref{fig:main}, which fixes the speaker and asks whether different listeners respond differently, we average each speaker's results over the L8B and L70B listeners, shown in \autoref{tab:avg_result}.
\method outperforms the speaker-oriented compression baseline Concise by a significant margin in Meeting Scheduling and MMLU-Pro-Debate, while remaining competitive in Text Pictionary. This suggests that an interruptible communication paradigm can improve the understanding level of different listeners to the same speaker: rather than passively receiving a compressed message that may not match their needs, the listener can actively signal when it has sufficient information. Speaker-oriented compression, by contrast, optimizes the message for a generic listener and cannot adapt to individual differences.

\begin{table}[!htp]\footnotesize\setlength{\tabcolsep}{4pt}
\centering
\caption{Fix the speakers and use different listeners. Average results across three speaker agents (\code{Llama-70B}, \code{Llama-405B}, and \code{Gemini-2.0-flash}).
``SR'' $=$ Success Rate. ``Generic'' and ``Concise'' are non-interruptible baselines; others are interruptible.}
\label{tab:avg_result}
\begin{tabular}{llccccc}
\toprule
Task & Metric & Generic & Concise & Random & Prompting & \method \\
\midrule
\multirow{2}{*}{Text Pictionary}
    & SR   & $0.758_{\pm 0.028}$ & $\mathbf{0.777}_{\pm 0.025}$ & $0.748_{\pm 0.033}$ & $0.582_{\pm 0.037}$ & $0.767_{\pm 0.022}$ \\
    & Cost & $373.6_{\pm 41.9}$ & $301.7_{\pm 30.9}$ & $389.9_{\pm 26.7}$ & $492.1_{\pm 34.1}$ & $\mathbf{278.1}_{\pm 37.8}$ \\
\midrule
\multirow{2}{*}{Meeting Scheduling}
    & SR   & $0.347_{\pm 0.059}$ & $0.347_{\pm 0.048}$ & $0.307_{\pm 0.060}$ & $0.310_{\pm 0.040}$ & $\mathbf{0.377}_{\pm 0.056}$ \\
    & Cost & $1294.2_{\pm 69.0}$ & $1090.1_{\pm 80.9}$ & $1672.5_{\pm 107.7}$ & $1781.6_{\pm 133.0}$ & $\mathbf{1026.5}_{\pm 85.0}$ \\
\midrule
\multirow{2}{*}{MMLU-Pro-Debate}
    & SR   & $0.583_{\pm 0.032}$ & $0.590_{\pm 0.074}$ & $0.580_{\pm 0.047}$ & $0.595_{\pm 0.068}$ & $\mathbf{0.620}_{\pm 0.075}$ \\
    & Cost & $1598.2_{\pm 101.6}$ & $838.1_{\pm 144.5}$ & $1714.1_{\pm 118.4}$ & $1952.8_{\pm 159.7}$ & $\mathbf{794.4}_{\pm 150.3}$ \\
\bottomrule
\end{tabular}
\end{table}

\begin{table*}[htbp]\centering\setlength{\tabcolsep}{4pt}
\small
\caption{Performance when we apply \method trained on one task (row) to other tasks (column). These experiments are based on \code{Llama-8B}. MMLU-D refer to MMLU-Pro-Debate}
\label{tab:cross}
\begin{tabular}{lccc|ccc}
\toprule
&\multicolumn{3}{c}{Success Ratio} &\multicolumn{3}{c}{Communication Cost} \\
\midrule
&TP &MS & MMLU-D &TP &MS & MMLU-D \\
\midrule
Generic & $0.743_{\pm 0.033}$ & $0.273_{\pm 0.077}$ & $0.520_{\pm 0.050}$ & $329_{\pm 37}$ & $1361_{\pm 86}$ & $1531_{\pm 126}$ \\
\midrule
Textual Pictionary &$0.743_{\pm 0.023}$ & $0.283_{\pm 0.640}$ & $0.537_{\pm 0.030}$ & $253_{\pm 27}$ & $1508_{\pm 84}$ & $879_{\pm 127}$ \\
Meeting Schedule &$0.726_{\pm 0.023}$ & $0.307_{\pm 0.049}$ & $0.543_{\pm 0.033}$ & $261_{\pm 31}$ & $1042_{\pm 83}$ & $905_{\pm 163}$ \\
MMLU-Pro-Debate &$0.726_{\pm 0.023}$ & $0.273_{\pm 0.540}$ & $0.583_{\pm 0.080}$ & $256_{\pm 28}$ & $1274_{\pm 83}$ & $782_{\pm 181}$ \\
\bottomrule
\end{tabular}
\end{table*}

\textbf{This interruption behavior can be generalized to other tasks.}
We also investigate whether interruption behaviors transfer across tasks by evaluating a model trained on one task on the remaining tasks. As shown in \autoref{tab:cross}, \method trained and evaluated on the same task performs best, which is expected since optimal interruption timing is task-specific. Nevertheless, the learned behavior still improves task performance and communication cost when transferred to new tasks.

\begin{figure}[htbp]
  \centering
  \begin{minipage}[c]{0.45\textwidth}
    \centering
    \includegraphics[width=0.8\linewidth]{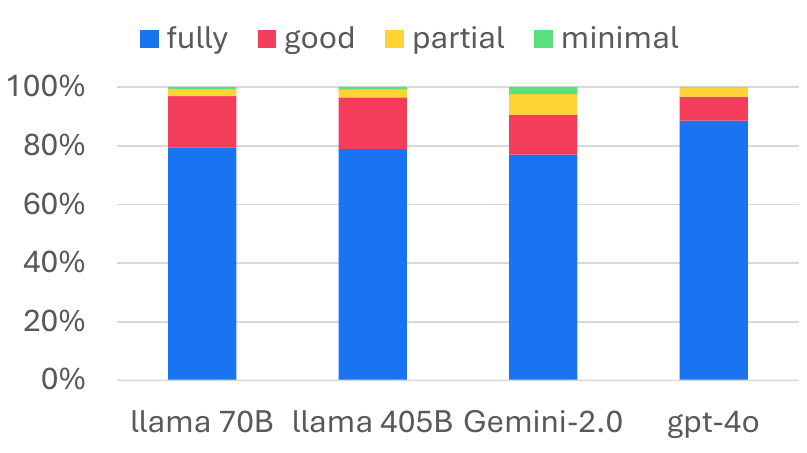}
    \captionof{figure}{\code{GPT-4o}'s understanding on incorrect guesses.}
    \label{fig:un_gpt_incorrect}
  \end{minipage}
  \hfill
  \begin{minipage}[c]{0.5\textwidth}
    \centering
    \captionof{table}{Meeting Scheduling with 3 interruptible agents. All three agents are based on the \code{Llama-70B} model.}
    \footnotesize\setlength{\tabcolsep}{2pt}
    \begin{tabular}{lcccc}
    \toprule
       & \multicolumn{1}{l}{Generic} & \multicolumn{1}{l}{Random} & \multicolumn{1}{l}{Prompting} & \multicolumn{1}{l}{\method} \\
    \midrule
    SR & $0.45_{\pm 0.04}$ & $0.45_{\pm 0.06}$ & $0.46_{\pm 0.09}$ & $0.45_{\pm 0.04}$ \\
    Cost & $1339_{\pm 26}$ & $1296_{\pm 74}$ & $1339_{\pm 11}$ & $1209_{\pm 41}$\\
    \bottomrule
    \end{tabular}
    \label{tab:complex}
  \end{minipage}
\end{figure}

\subsection{More Complex Interruption Patterns}
\label{sec:complex-interruption}
As a first attempt in listener-oriented communication, this paper focuses on the basic communication pattern where only one agent can interrupt the other agents. However, as we mentioned in Section \ref{sec:protocol}, our interruptible communication mechanism can also be applied to other multi-agent communication scenarios. 

We extend Meeting Scheduling to symmetric interruptible communication, allowing all three agents (2 planner agents and 1 traveler agent) to interrupt each other in a cascading manner. We directly apply the 70B \method to this setting without additional fine-tuning. 

From the results in \autoref{tab:complex}, our \method saves the communication cost without performance degradation. This shows the promise of our method in genuinely open, multi-directional multi-agent environments. However, we also find that while the \method’s interruption reduces the per-round cost by half (103 $\rightarrow$ 64 tokens), it also improves the number of communication rounds (13.39$\rightarrow$20.29 rounds). Further investigation, such as additional finetuning, can be made to improve the interruption efficiency in these complex scenarios, showcasing the extensibility of our proposed framework.

\section{Related Work}
\textbf{Multi-agent LLM Framework}
Studies have shown that debating with other agents can encourage the exchange of various solutions, thus enhancing LLMs' capabilities in reasoning~\citep{du2023improving,liang2023encouraging,yin2023exchange} and evaluation~\citep{chern2024can,chan2024chateval}. Meanwhile, decomposing complex tasks into subtasks and assigning them to different agents can also improve problem-solving through collaboration~\citep{liu2023dynamic,ChenDSZS00024}. Recent studies further explore multi-agent design optimization~\cite{tran2025multi}, such as better prompts and topologies~\citep{zhou2025multi,zhuge2024language} and investigate scaling principles for effective large-scale multi-agent collaboration~\citep{qian2025scaling}. These approaches collectively optimize multi-agent frameworks toward more effective collaboration, while often ignoring communication efficiency.

\textbf{Communication Topology in Multi-agent Framework} 
Recent research has focused on optimizing agent interactions to reduce token overhead. \citet{li-etal-2024-improving-multi} demonstrate that sparse communication topologies can effectively improve multi-agent debate performance while reducing unnecessary messages. AgentPrune~\citep{zhang2025cut} performs one-shot pruning on spatial-temporal message-passing graphs to eliminate redundant communications. AgentDropout~\citep{wang-etal-2025-agentdropout} dynamically eliminates redundant agents and optimizes adjacency matrices across communication rounds.
These methods reduce unnecessary communication between agents to improve system efficiency; however, the communication cost within individual message exchanges remains high.

\textbf{Compression for LLM Efficiency}
For compressing reasoning outputs, TokenSkip~\citep{xia2025tokenskip} and C3oT~\citep{kang2025c3ot} reduce chain-of-thought length at the token level, while CODI~\citep{lee2023codi} and Coconut~\citep{hao2025training} compress reasoning into a continuous latent space. A related line of work applies summarization directly within multi-agent pipelines: Chain-of-Agents~\citep{zhang2024chain} uses sequential inter-agent summarization for long-context handling, and dedicated summarizer agents~\citep{yan2025beyond} have been proposed as a coordination mechanism to reduce message volume. However, they compress from the speaker side, making it difficult to identify the parts that are indispensable to various listeners.

\section{Conclusion}
We propose an interruptible communication framework \method for multi-agent systems. \method allows the agent to decide when to interrupt, which reduces generation cost and communication latency. 
We calibrate agents' estimation of suitable interruption with high task rewards and low communication costs. Experiments on multiple \speaker-\listener settings under 3 multi-agent scenarios show that \method can reduce communication costs while achieving comparable or higher task performance. Further investigation shows that it can generalize to different tasks and \speakers.

\bibliography{verified}

\begin{thebibliography}{51}
\providecommand{\natexlab}[1]{#1}
\providecommand{\url}[1]{\texttt{#1}}
\expandafter\ifx\csname urlstyle\endcsname\relax
  \providecommand{\doi}[1]{doi: #1}\else
  \providecommand{\doi}{doi: \begingroup \urlstyle{rm}\Url}\fi

\bibitem[{Anthropic}(2025)]{anthropic2025agent}
{Anthropic}.
\newblock https://www.anthropic.com/engineering/multi-agent-research-system.
\newblock \url{https://www.anthropic.com/engineering/multi-agent-research-system}, 2025.

\bibitem[Barres et~al.()Barres, Dong, Ray, Si, and Narasimhan]{barres2025tau2}
Victor Barres, Honghua Dong, Soham Ray, Xujie Si, and Karthik Narasimhan.
\newblock $\tau^2$-bench: Evaluating conversational agents in a dual-control environment.
\newblock URL \url{https://arxiv.org/abs/2506.07982}.

\bibitem[Bennett(1978)]{bennett1978interruptions}
Adrian Bennett.
\newblock Interruptions and the interpretation of conversation.
\newblock In \emph{Annual Meeting of the Berkeley Linguistics Society}, volume~4, pp.\  557. Linguistic Society of America, 1978.
\newblock \doi{10.3765/BLS.V4I0.2199}.

\bibitem[Cemri et~al.(2025)Cemri, Pan, Yang, Agrawal, Chopra, Tiwari, Keutzer, Parameswaran, Klein, Ramchandran, et~al.]{cemri2025multi}
M.~Cemri, Melissa~Z. Pan, Shuyi Yang, Lakshya~A. Agrawal, Bhavya Chopra, Rishabh Tiwari, Kurt Keutzer, Aditya~G. Parameswaran, Dan Klein, K.~Ramchandran, et~al.
\newblock Why do multi-agent llm systems fail?
\newblock \emph{Advances in Neural Information Processing Systems 38}, pp.\  135676--135729, 2025.
\newblock \doi{10.52202/085713-4082}.

\bibitem[Chan et~al.(2024)Chan, Chen, Su, Yu, Xue, Zhang, Fu, and Liu]{chan2024chateval}
Chi-Min Chan, Weize Chen, Yusheng Su, Jianxuan Yu, Wei Xue, Shanghang Zhang, Jie Fu, and Zhiyuan Liu.
\newblock Chateval: Towards better {LLM}-based evaluators through multi-agent debate.
\newblock In \emph{arXiv.org}, 2024.
\newblock \doi{10.48550/arXiv.2308.07201}.
\newblock URL \url{https://openreview.net/forum?id=FQepisCUWu}.

\bibitem[Chen et~al.(2024)Chen, Dong, Shu, Zhang, Sesay, Karlsson, Fu, and Shi]{ChenDSZS00024}
Guangyao Chen, Siwei Dong, Yu~Shu, Ge~Zhang, Jaward Sesay, Börje Karlsson, Jie Fu, and Yemin Shi.
\newblock Autoagents: A framework for automatic agent generation.
\newblock In \emph{International Joint Conference on Artificial Intelligence}, pp.\  22--30, 2024.
\newblock \doi{10.48550/arXiv.2309.17288}.
\newblock URL \url{https://www.ijcai.org/proceedings/2024/3}.

\bibitem[Chern et~al.(2024)Chern, Chern, Neubig, and Liu]{chern2024can}
Steffi Chern, Ethan Chern, Graham Neubig, and Pengfei Liu.
\newblock Can large language models be trusted for evaluation? scalable meta-evaluation of llms as evaluators via agent debate.
\newblock \emph{arXiv.org}, 2024.
\newblock \doi{10.48550/arXiv.2401.16788}.

\bibitem[Crawford \& Veloso(2004)Crawford and Veloso]{crawford2006mechanism}
Elisabeth Crawford and Manuela Veloso.
\newblock Mechanism design for multi-agent meeting scheduling.
\newblock \emph{ACM International Conference on International Agent Technology}, 4\penalty0 (2):\penalty0 209--220, 2004.
\newblock \doi{10.3233/WEB-2006-wia00087}.

\bibitem[Du et~al.(2023)Du, Li, Torralba, Tenenbaum, and Mordatch]{du2023improving}
Yilun Du, Shuang Li, Antonio Torralba, Joshua~B Tenenbaum, and Igor Mordatch.
\newblock Improving factuality and reasoning in language models through multiagent debate.
\newblock \emph{International Conference on Machine Learning}, 2023.
\newblock \doi{10.48550/arXiv.2305.14325}.

\bibitem[Dubey et~al.(2024)Dubey, Jauhri, Pandey, Kadian, Al-Dahle, Letman, Mathur, Schelten, Yang, Fan, et~al.]{grattafiori2024llama}
Abhimanyu Dubey, Abhinav Jauhri, Abhinav Pandey, Abhishek Kadian, Ahmad Al-Dahle, Aiesha Letman, Akhil Mathur, A.~Schelten, Amy Yang, Angela Fan, et~al.
\newblock The llama 3 herd of models.
\newblock \emph{arXiv}, 2024.

\bibitem[Dubois et~al.(2023)Dubois, Li, Taori, Zhang, Gulrajani, Ba, Guestrin, Liang, and Hashimoto]{dubois2023alpacafarm}
Yann Dubois, Chen~Xuechen Li, Rohan Taori, Tianyi Zhang, Ishaan Gulrajani, Jimmy Ba, Carlos Guestrin, Percy~S Liang, and Tatsunori~B Hashimoto.
\newblock Alpacafarm: A simulation framework for methods that learn from human feedback.
\newblock \emph{Advances in Neural Information Processing Systems 36}, 36:\penalty0 30039--30069, 2023.
\newblock \doi{10.52202/075280-1308}.

\bibitem[Fang et~al.(2025)Fang, Ma, and Wang]{fang2025thinkless}
Gongfan Fang, Xinyin Ma, and Xinchao Wang.
\newblock Thinkless: Llm learns when to think.
\newblock \emph{Advances in Neural Information Processing Systems 38}, pp.\  167994--168021, 2025.
\newblock \doi{10.52202/085713-5060}.

\bibitem[Guo et~al.(2024)Guo, Chen, Wang, Chang, Pei, Chawla, Wiest, and Zhang]{guo2024large}
Taicheng Guo, Xiuying Chen, Yaqi Wang, Ruidi Chang, Shichao Pei, Nitesh~V Chawla, Olaf Wiest, and Xiangliang Zhang.
\newblock Large language model based multi-agents: A survey of progress and challenges.
\newblock In \emph{International Joint Conference on Artificial Intelligence}, 2024.
\newblock \doi{10.48550/arXiv.2402.01680}.

\bibitem[Hao et~al.(2025)Hao, Sukhbaatar, Su, Li, Hu, Weston, and Tian]{hao2025training}
Shibo Hao, Sainbayar Sukhbaatar, DiJia Su, Xian Li, Zhiting Hu, Jason~E Weston, and Yuandong Tian.
\newblock Training large language models to reason in a continuous latent space.
\newblock In \emph{arXiv.org}, 2025.
\newblock \doi{10.48550/arXiv.2412.06769}.
\newblock URL \url{https://openreview.net/forum?id=Itxz7S4Ip3}.

\bibitem[He et~al.(2024)He, Treude, and Lo]{he2025llm}
Junda He, Christoph Treude, and David Lo.
\newblock Llm-based multi-agent systems for software engineering: Literature review, vision, and the road ahead.
\newblock \emph{ACM Transactions on Software Engineering and Methodology}, 34\penalty0 (5):\penalty0 1--30, 2024.
\newblock \doi{10.1145/3712003}.

\bibitem[Hu et~al.(2025)Hu, Cai, Du, Zhu, Liu, Yu, Hou, Tang, and Chen]{hu2025selfevolving}
Yue Hu, Yuzhu Cai, Yaxin Du, Xinyu Zhu, Xiangrui Liu, Zijie Yu, Yuchen Hou, Shuo Tang, and Siheng Chen.
\newblock Self-evolving multi-agent collaboration networks for software development.
\newblock In \emph{International Conference on Learning Representations}, 2025.
\newblock \doi{10.48550/arXiv.2410.16946}.
\newblock URL \url{https://openreview.net/forum?id=4R71pdPBZp}.

\bibitem[Hurst et~al.(2024)Hurst, Lerer, Goucher, Perelman, Ramesh, Clark, Ostrow, Welihinda, Hayes, Radford, et~al.]{hurst2024gpt}
OpenAI~Aaron Hurst, Adam Lerer, Adam~P. Goucher, Adam Perelman, Aditya Ramesh, Aidan Clark, AJ~Ostrow, Akila Welihinda, Alan Hayes, Alec Radford, et~al.
\newblock Gpt-4o system card.
\newblock \emph{arXiv}, 2024.

\bibitem[Kang et~al.(2025)Kang, Sun, Chen, and Zou]{kang2025c3ot}
Yu~Kang, Xianghui Sun, Liangyu Chen, and Wei Zou.
\newblock C3ot: Generating shorter chain-of-thought without compromising effectiveness.
\newblock In \emph{Proceedings of the AAAI Conference on Artificial Intelligence}, volume~39, pp.\  24312--24320. Association for the Advancement of Artificial Intelligence (AAAI), 2025.
\newblock \doi{10.1609/aaai.v39i23.34608}.

\bibitem[Khan et~al.(2024)Khan, Hughes, Valentine, Ruis, Sachan, Radhakrishnan, Grefenstette, Bowman, Rocktaschel, and Perez]{khan2024debating}
Akbir Khan, John Hughes, Dan Valentine, L.~Ruis, Kshitij Sachan, Ansh Radhakrishnan, Edward Grefenstette, Samuel~R. Bowman, Tim Rocktaschel, and Ethan Perez.
\newblock Debating with more persuasive llms leads to more truthful answers.
\newblock \emph{International Conference on Machine Learning}, 235:\penalty0 23662--23733, 2024.
\newblock \doi{10.48550/arXiv.2402.06782}.

\bibitem[Lee et~al.(2023)Lee, Kim, and Park]{lee2023codi}
Chaejeong Lee, Jayoung Kim, and Noseong Park.
\newblock Codi: Co-evolving contrastive diffusion models for mixed-type tabular synthesis.
\newblock In \emph{International Conference on Machine Learning}, pp.\  18940--18956. PMLR, 2023.
\newblock \doi{10.48550/arXiv.2304.12654}.

\bibitem[Li et~al.(2024)Li, Du, Zhang, Hou, Grabowski, Li, and Ie]{li-etal-2024-improving-multi}
Yunxuan Li, Yibing Du, Jiageng Zhang, Le~Hou, Peter Grabowski, Yeqing Li, and Eugene Ie.
\newblock Improving multi-agent debate with sparse communication topology.
\newblock In Yaser Al-Onaizan, Mohit Bansal, and Yun-Nung Chen (eds.), \emph{Findings of the Association for Computational Linguistics: EMNLP 2024}, pp.\  7281--7294. Association for Computational Linguistics, 2024.
\newblock \doi{10.18653/v1/2024.findings-emnlp.427}.

\bibitem[Liang et~al.(2024)Liang, He, Jiao, Wang, Wang, Wang, Yang, Tu, and Shi]{liang2023encouraging}
Tian Liang, Zhiwei He, Wenxiang Jiao, Xing Wang, Yan Wang, Rui Wang, Yujiu Yang, Zhaopeng Tu, and Shuming Shi.
\newblock Encouraging divergent thinking in large language models through multi-agent debate.
\newblock \emph{Proceedings of the 2024 Conference on Empirical Methods in Natural Language Processing}, pp.\  17889--17904, 2024.
\newblock \doi{10.18653/v1/2024.emnlp-main.992}.

\bibitem[Liu et~al.(2024)Liu, Zhang, Li, Liu, and Yang]{liu2023dynamic}
Zijun Liu, Yanzhe Zhang, Peng Li, Yang Liu, and Diyi Yang.
\newblock Dynamic llm-agent network: An llm-agent collaboration framework with agent team optimization.
\newblock In \emph{COLM}, 2024.

\bibitem[Lycan(1977)]{lycan1977conversation}
William~G Lycan.
\newblock Conversation, politeness, and interruption.
\newblock \emph{Paper in Linguistics}, 10\penalty0 (1-2):\penalty0 23--53, 1977.
\newblock \doi{10.1080/08351819709370438}.

\bibitem[Ng et~al.(1995)Ng, Brooke, and Dunne]{ng1995interruption}
Sik~Hung Ng, Mark Brooke, and Michael Dunne.
\newblock Interruption and influence in discussion groups.
\newblock \emph{Journal of Language and Social Psychology}, 14\penalty0 (4):\penalty0 369--381, 1995.
\newblock \doi{10.1177/0261927X950144003}.

\bibitem[{OpenClaw}(2026)]{openclaw2026}
{OpenClaw}.
\newblock Openclaw.
\newblock \url{https://github.com/openclaw/openclaw}, 2026.

\bibitem[Park et~al.(2023)Park, O'Brien, Cai, Morris, Liang, and Bernstein]{park2023generative}
Joon~Sung Park, Joseph O'Brien, Carrie~Jun Cai, Meredith~Ringel Morris, Percy Liang, and Michael~S Bernstein.
\newblock Generative agents: Interactive simulacra of human behavior.
\newblock In \emph{ACM Symposium on User Interface Software and Technology}, pp.\  1--22. ACM, 2023.
\newblock \doi{10.1145/3586183.3606763}.

\bibitem[Qian et~al.(2025{\natexlab{a}})Qian, Liu, Prabhakar, Liu, Zhang, Chen, Ji, Yao, Heinecke, Savarese, et~al.]{qian2025userbench}
Cheng Qian, Zuxin Liu, Akshara Prabhakar, Zhiwei Liu, Jianguo Zhang, Haolin Chen, Heng Ji, Weiran Yao, Shelby Heinecke, Silvio Savarese, et~al.
\newblock Userbench: An interactive gym environment for user-centric agents.
\newblock \emph{arXiv preprint arXiv:2507.22034}, 2025{\natexlab{a}}.

\bibitem[Qian et~al.(2025{\natexlab{b}})Qian, Xie, Wang, Liu, Dang, Du, Chen, Yang, Liu, and Sun]{qian2025scaling}
Cheng Qian, Zihao Xie, Yifei Wang, Wei Liu, Yufan Dang, Zhuoyun Du, Weize Chen, Cheng Yang, Zhiyuan Liu, and Maosong Sun.
\newblock Scaling large language model-based multi-agent collaboration.
\newblock In \emph{International Conference on Learning Representations}, 2025{\natexlab{b}}.
\newblock \doi{10.48550/arXiv.2406.07155}.
\newblock URL \url{https://openreview.net/forum?id=K3n5jPkrU6}.

\bibitem[Qiao et~al.(2025)Qiao, Deng, Zeng, Wang, Wei, Meng, Zhou, Ren, and Zhang]{qiao2025concise}
Ziqing Qiao, Yongheng Deng, Jiali Zeng, Dong Wang, Lai Wei, Fandong Meng, Jie Zhou, Ju~Ren, and Yaoxue Zhang.
\newblock Concise: Confidence-guided compression in step-by-step efficient reasoning.
\newblock \emph{Proceedings of the 2025 Conference on Empirical Methods in Natural Language Processing}, pp.\  8021--8040, 2025.
\newblock \doi{10.18653/v1/2025.emnlp-main.405}.

\bibitem[Renting et~al.(2024)Renting, Hoos, and Jonker]{renting2024multi}
Bram~M Renting, Holger Hoos, and Catholijn~M Jonker.
\newblock Multi-agent meeting scheduling: A negotiation perspective.
\newblock In \emph{The Sixteenth Workshop on Adaptive and Learning Agents}, 2024.

\bibitem[Seshadri et~al.()Seshadri, Cahyawijaya, Odumakinde, Singh, and Goldfarb-Tarrant]{seshadri2026lost}
Preethi Seshadri, Samuel Cahyawijaya, Ayomide Odumakinde, Sameer Singh, and Seraphina Goldfarb-Tarrant.
\newblock Lost in simulation: Llm-simulated users are unreliable proxies for human users in agentic evaluations.

\bibitem[Shannon et~al.(1959)]{shannon1959coding}
Claude~E Shannon et~al.
\newblock Coding theorems for a discrete source with a fidelity criterion.
\newblock \emph{IRE Nat. Conv. Rec}, 4\penalty0 (142-163):\penalty0 1, 1959.

\bibitem[Team et~al.(2023)Team, Anil, Borgeaud, Alayrac, Yu, Soricut, Schalkwyk, Dai, Hauth, Millican, et~al.]{team2023gemini}
Gemini Team, Rohan Anil, Sebastian Borgeaud, Jean-Baptiste Alayrac, Jiahui Yu, Radu Soricut, Johan Schalkwyk, Andrew~M Dai, Anja Hauth, Katie Millican, et~al.
\newblock Gemini: a family of highly capable multimodal models.
\newblock \emph{arXiv preprint arXiv:2312.11805}, 2023.

\bibitem[Team et~al.(2026)Team, Bai, Bai, Bao, Cai, Cao, Charles, Che, Chen, Chen, et~al.]{team2026kimi}
Kimi Team, Tongtong Bai, Yifan Bai, Yiping Bao, SH~Cai, Yuan Cao, Y~Charles, HS~Che, Cheng Chen, Guanduo Chen, et~al.
\newblock Kimi k2. 5: Visual agentic intelligence.
\newblock \emph{arXiv preprint arXiv:2602.02276}, 2026.

\bibitem[Tran et~al.(2025)Tran, Dao, Nguyen, Pham, O'Sullivan, and Nguyen]{tran2025multi}
Khanh-Tung Tran, Dung Dao, Minh-Duong Nguyen, Quoc-Viet Pham, Barry O'Sullivan, and Hoang~D Nguyen.
\newblock Multi-agent collaboration mechanisms: A survey of llms.
\newblock \emph{arXiv.org}, 2025.
\newblock \doi{10.48550/arXiv.2501.06322}.

\bibitem[Veluri et~al.(2024)Veluri, Peloquin, Yu, Gong, and Gollakota]{veluri2024beyond}
Bandhav Veluri, Benjamin~N Peloquin, Bokai Yu, Hongyu Gong, and Shyamnath Gollakota.
\newblock Beyond turn-based interfaces: Synchronous llms as full-duplex dialogue agents.
\newblock In \emph{Proceedings of the 2024 Conference on Empirical Methods in Natural Language Processing}, pp.\  21390--21402. Association for Computational Linguistics, 2024.
\newblock \doi{10.18653/v1/2024.emnlp-main.1192}.

\bibitem[Wang et~al.(2023)Wang, Liu, Zheng, Qi, Chen, Yang, Zhao, Wang, Song, and Huang]{Wang2023AvalonsGO}
Shenzhi Wang, Chang Liu, Zilong Zheng, Siyuan Qi, Shuo Chen, Qisen Yang, Andrew Zhao, Chaofei Wang, Shiji Song, and Gao Huang.
\newblock Avalon's game of thoughts: Battle against deception through recursive contemplation.
\newblock \emph{arXiv.org}, abs/2310.01320, 2023.
\newblock \doi{10.48550/arXiv.2310.01320}.
\newblock URL \url{https://api.semanticscholar.org/CorpusID:263605971}.

\bibitem[Wang et~al.(2024)Wang, Ma, Zhang, Ni, Chandra, Guo, Ren, Arulraj, He, Jiang, et~al.]{wang2024mmlu}
Yubo Wang, Xueguang Ma, Ge~Zhang, Yuansheng Ni, Abhranil Chandra, Shiguang Guo, Weiming Ren, Aaran Arulraj, Xuan He, Ziyan Jiang, et~al.
\newblock Mmlu-pro: A more robust and challenging multi-task language understanding benchmark.
\newblock \emph{Advances in Neural Information Processing Systems 37}, 37:\penalty0 95266--95290, 2024.
\newblock \doi{10.52202/079017-3018}.

\bibitem[Wang et~al.(2025)Wang, Wang, Liu, Ding, Zhang, Liu, and Zhang]{wang-etal-2025-agentdropout}
Zhexuan Wang, Yutong Wang, Xuebo Liu, Liang Ding, Miao Zhang, Jie Liu, and Min Zhang.
\newblock {A}gent{D}ropout: Dynamic agent elimination for token-efficient and high-performance {LLM}-based multi-agent collaboration.
\newblock In Wanxiang Che, Joyce Nabende, Ekaterina Shutova, and Mohammad~Taher Pilehvar (eds.), \emph{Proceedings of the 63rd Annual Meeting of the Association for Computational Linguistics (Volume 1: Long Papers)}, pp.\  24013--24035. Association for Computational Linguistics, 2025.
\newblock \doi{10.18653/v1/2025.acl-long.1170}.

\bibitem[Xia et~al.(2025)Xia, Leong, Wang, Li, and Li]{xia2025tokenskip}
Heming Xia, Chak~Tou Leong, Wenjie Wang, Yongqi Li, and Wenjie Li.
\newblock Tokenskip: Controllable chain-of-thought compression in llms.
\newblock In \emph{Proceedings of the 2025 Conference on Empirical Methods in Natural Language Processing}, pp.\  3351--3363. Association for Computational Linguistics, 2025.
\newblock \doi{10.18653/v1/2025.emnlp-main.165}.

\bibitem[Xu et~al.(2023)Xu, Wang, Li, Luo, Wang, Liu, and Liu]{xu2023exploring}
Yuzhuang Xu, Shuo Wang, Peng Li, Fuwen Luo, Xiaolong Wang, Weidong Liu, and Yang Liu.
\newblock Exploring large language models for communication games: An empirical study on werewolf.
\newblock \emph{arXiv.org}, 2023.
\newblock \doi{10.48550/arXiv.2309.04658}.

\bibitem[Yan et~al.(2025)Yan, Zhou, Zhang, Zhang, Zhou, Miao, Li, Li, and Zhang]{yan2025beyond}
Bingyu Yan, Zhibo Zhou, Litian Zhang, Lian Zhang, Ziyi Zhou, Dezhuang Miao, Zhoujun Li, Chaozhuo Li, and Xiaoming Zhang.
\newblock Beyond self-talk: A communication-centric survey of llm-based multi-agent systems.
\newblock \emph{arXiv preprint arXiv:2502.14321}, 2025.

\bibitem[Yao et~al.()Yao, Shinn, Razavi, and Narasimhan]{yao2025taubench}
Shunyu Yao, Noah Shinn, Pedram Razavi, and Karthik~R Narasimhan.
\newblock \{\${\textbackslash}tau\$\}-bench: A benchmark for {\textbackslash}underline\{T\}ool-{\textbackslash}underline\{A\}gent-{\textbackslash}underline\{U\}ser interaction in real-world domains.
\newblock In \emph{The Thirteenth International Conference on Learning Representations}.
\newblock URL \url{https://openreview.net/forum?id=roNSXZpUDN}.

\bibitem[Yin et~al.(2023)Yin, Sun, Chang, Guo, Dai, Huang, and Qiu]{yin2023exchange}
Zhangyue Yin, Qiushi Sun, Cheng Chang, Qipeng Guo, Junqi Dai, Xuan-Jing Huang, and Xipeng Qiu.
\newblock Exchange-of-thought: Enhancing large language model capabilities through cross-model communication.
\newblock In \emph{Proceedings of the 2023 Conference on Empirical Methods in Natural Language Processing}, pp.\  15135--15153. Association for Computational Linguistics, 2023.
\newblock \doi{10.18653/v1/2023.emnlp-main.936}.

\bibitem[Zhang et~al.(2025)Zhang, Yue, Li, Yun, Wan, Wang, Cheng, Yu, and Chen]{zhang2025cut}
Guibin Zhang, Yanwei Yue, Zhixun Li, Sukwon Yun, Guancheng Wan, Kun Wang, Dawei Cheng, Jeffrey~Xu Yu, and Tianlong Chen.
\newblock Cut the crap: An economical communication pipeline for {LLM}-based multi-agent systems.
\newblock In \emph{International Conference on Learning Representations}, 2025.
\newblock \doi{10.48550/arXiv.2410.02506}.
\newblock URL \url{https://openreview.net/forum?id=LkzuPorQ5L}.

\bibitem[Zhang et~al.(2024{\natexlab{a}})Zhang, Chen, Hu, Han, Xu, Xu, Zhao, Sun, and Liu]{zhang2024beyond}
Xinrong Zhang, Yingfa Chen, Shengding Hu, Xu~Han, Zihang Xu, Yuanwei Xu, Weilin Zhao, Maosong Sun, and Zhiyuan Liu.
\newblock Beyond the turn-based game: Enabling real-time conversations with duplex models.
\newblock In \emph{Proceedings of the 2024 Conference on Empirical Methods in Natural Language Processing}, pp.\  11543--11557. Association for Computational Linguistics, 2024{\natexlab{a}}.
\newblock \doi{10.18653/v1/2024.emnlp-main.644}.

\bibitem[Zhang et~al.(2024{\natexlab{b}})Zhang, Sun, Chen, Pfister, Zhang, and Arik]{zhang2024chain}
Yusen Zhang, Ruoxi Sun, Yanfei Chen, Tomas Pfister, Rui Zhang, and Sercan~Ö. Arik.
\newblock Chain of agents: Large language models collaborating on long-context tasks.
\newblock \emph{Advances in Neural Information Processing Systems 37}, 37:\penalty0 132208--132237, 2024{\natexlab{b}}.
\newblock \doi{10.52202/079017-4202}.

\bibitem[Zheng et~al.(2024)Zheng, Mishra, Zhang, Chen, Chen, Nova, Hou, Cheng, Le, Chi, et~al.]{zheng2024natural}
Huaixiu~Steven Zheng, Swaroop Mishra, Hugh Zhang, Xinyun Chen, Minmin Chen, Azade Nova, Le~Hou, Heng-tze Cheng, Quoc~V. Le, E.~Chi, et~al.
\newblock Natural plan: Benchmarking llms on natural language planning.
\newblock \emph{arXiv.org}, 2024.
\newblock \doi{10.48550/arXiv.2406.04520}.

\bibitem[Zhou et~al.(2025)Zhou, Wan, Wan, Sun, Palangi, Iqbal, Vuli'c, Korhonen, and Arik]{zhou2025multi}
Han Zhou, Xingchen Wan, Xingchen Wan, Ruoxi Sun, Hamid Palangi, Shariq Iqbal, Ivan Vuli'c, Anna Korhonen, and Sercan~Ö. Arik.
\newblock Multi-agent design: Optimizing agents with better prompts and topologies.
\newblock \emph{arXiv.org}, 2025.
\newblock \doi{10.48550/arXiv.2502.02533}.

\bibitem[Zhuge et~al.(2024)Zhuge, Wang, Kirsch, Faccio, Khizbullin, and Schmidhuber]{zhuge2024language}
Mingchen Zhuge, Wenyi Wang, Louis Kirsch, Francesco Faccio, Dmitrii Khizbullin, and J{\"u}rgen Schmidhuber.
\newblock Language agents as optimizable graphs.
\newblock \emph{International Conference on Machine Learning}, 2024.
\newblock \doi{10.48550/arXiv.2402.16823}.

\end{thebibliography}
\bibliographystyle{colm2026_conference}

\newpage
\appendix
\section{Task Setup}
\label{sec:app-task-setup}
Our task selection follows three principles: (i) communication-dependence, where the task relies on inter-agent communication; (ii) necessity of multi-agent setup, such as multi-player games and agents with private information; and (iii) coverage of common multi-agent tasks, such as debate for reasoning.

Based on these principles, Text Pictionary reflects a real-world two-player communication game; meeting scheduling simulates a practical daily task where agents hold private information that cannot be shared directly~\citep{crawford2006mechanism,renting2024multi}; and multi-agent debate is widely used in reasoning tasks for code and math~\citep{du2023improving,liang2023encouraging,li-etal-2024-improving-multi,khan2024debating}. We will clarify these points in the revision.

We describe the setup of our three tasks and leave the task-specific prompts for each agent in \autoref{sec:app-task-prompt} and data statistics in Table \ref{tab:ds}.

\begin{table}[!htp]\footnotesize\setlength{\tabcolsep}{4pt}
\centering
\caption{Dataset Statistics. Inter and No-inter indicate the percentage of interruption and no interruption in the training trajectories. Avg. \#Message is the average number of messages in the training trajectories, and Avg Length is the average length of the trajectory in words.}
\label{tab:ds}
\begin{tabular}{lcccccccc}
\toprule
&Train &Test & \# Tractories & Inter & No-inter & Avg. \#Message & Avg Length \\
\midrule
Textual Pictionary &112 &100 &3010 &54.1\% &45.9\% &5.1 &425.4 \\
Meeting Schedule &100 &50 &1973 &62.0\% &38.0\% &14.7 &1182.7 \\
MMLU-Pro-Debate &200 &100 &4431 &53.6\% &46.4\% &5.1 &1448.1 \\
\bottomrule
\end{tabular}
\end{table}

\textbf{Text Pictionary}
We collect entities from online pictionary games and filter out some easy ones that can be successfully guessed within one round by Llama 3 8B. In each round, the describer provides a description of the given secret entity, and the guesser can provide its guess or ask clarification questions. We check the guesser's answer after each round and terminate the game once its answer matches the target entity. If the describer reveals the secret entity in any way, the game terminates immediately with a reward of 0.

\textbf{Meeting Scheduling}
We synthesize data with location, travel time, and meeting time constraints for the meeting scheduling task. There is one traveler and two planning agents working together to schedule the meeting. A planning agent helps a person who wants to meet with the traveler at their location. It has access to private information about this person, including location, available time slots, meeting duration, and preferred meeting times. The traveler needs to travel among locations to meet these two people separately. The final meeting schedule should satisfy the following hard constraints: \textit{(i) the location, duration, time, and participant of the scheduled meeting should match the requirements; (ii) the travel time between locations should fit within the free time between meetings; (iii) no extra meeting is scheduled.} Each task case includes the distance matrix and the private information for the planning agents and traveler. We create a script to randomly generate cases and verify them. During the discussion, each agent can choose the next agent it wants to chat with. The traveler can interrupt the planning agents' messages. The discussion stops once the traveler provides a meeting schedule or reaches the maximum round. If all three constraints are satisfied, the task receives a reward of 1.

\textbf{Multi-agent Debate for MMLU-Pro}
We adapt the original multiple-choice MMLU-Pro questions to a 3-agent setting. For each question, we let Llama 3 405B generate an explanation for the correct answer and provide the most confusing incorrect answer with explanation. Then we randomly assign the correct and incorrect answers to the Pro and Con side debaters. The two debaters need to argue for their assigned answer, and the moderator should vote for one side based on the discussion. In each round, the Pro and Con sides take turns providing their statements based on the assigned answer and explanation. At the end of the round, the moderator votes for one side or waits for another round. The debate stops when the moderator provides its vote or reaches the maximum round. The moderator can interrupt the Pro side and let the Con side start its statement directly. It can also interrupt the Con side and provide its preference.
If the moderator selects the correct answer, it receives a reward of 1.

Our MMLU-Pro Debate evaluation captures the same essential multi-round structure: agents with distinct assigned positions exchange arguments across turns, and a third agent decides when it has gathered enough information to reach a decision. 

By showing that \method reduces communication cost on debate-style reasoning tasks while improving success rate, we demonstrate its promise for the verbose multi-turn exchanges that characterize multi-agent coding systems.

\section{Decompose More Complex Multi-agent Communication}
\label{sec:app-discussion}
As we discussed in Section \ref{sec:complex-interruption}, more complex scenarios such as cascading or conflicting interruptions, or mutual communication can be handled in this way. Suppose there are three agents: Alice, Bob and Charlie, and we demonstrate the solutions as follows: 
\begin{itemize}[noitemsep, leftmargin=2em, topsep=1pt]
    \item \textbf{Communication with a fixed order } (Alice $\rightarrow$ Bob $\rightarrow$ Charlie $\rightarrow$… ): It can be decomposed into independent communication Alice $\rightarrow$ Bob, Bob $\rightarrow$ Charlie, Charlie $\rightarrow \cdots$
    \item \textbf{Free discussion}: (Alice $\rightarrow$ Bob while Alice $\rightarrow$ Charlie, Alice broadcasting to Bob and Charlie): Bob and Charlie independently decide the interruption, while Alice stops at the first interruption point to avoid conflicting or cascading. 
\end{itemize}

\textbf{Mutual or symmetric communication} (Alice $\rightarrow$ Bob $\rightarrow$ Alice $\rightarrow$ … ) is a special case of the fixed order. It can also be decomposed into multiple independent communication channels Alice $\rightarrow$ Bob, Bob $\rightarrow$ Alice, …, each process can use \method to decide interruption. 
These scenarios can be decomposed into the basic communication pattern we investigated in our paper, showing the great promise of \method. We would like to leave these as our future directions.

\section{Training and Evaluation Details}
\label{sec:app-training}
\subsection{Rollout setting}
We use Llama 3 70B, 405B, and Gemini models as the describers in text pictionary, planning agents in meeting scheduling, and debaters in multi-agent debate. We let them communicate with the Llama 8B and 70B agents and sample the trajectories as described in Appendix \ref{sec:learn}. During sampling, we use a chunk size of 8 for text pictionary and 16 for the other two tasks. We set a maximum branch size of 3 and a rollout number of 10, with a maximum of 10 rounds for each rollout. For one case, we choose at most 10 non-terminating speaking nodes. For each non-terminating speaking node, we roll out for all its child nodes, which are the chunks of the next agent's message. The path from the root to this node is the chat history, and the child node is the current partial message. We assign interruption labels and formulate the trajectories based on the tree. The temperature is set to 1 for sampling. 

\subsection{One-time computation cost for tree sampling.} 
\label{sec:app-training-cost}
More generally, suppose there are $n$ agents, and each agent takes its turn to speak. For each interruptible agent, we do a separate tree sampling, i.e., we assume only one agent can interrupt in one tree sampling. Therefore, in one round, there are $(n-1)$ messages to be interrupted and create $B$ interrupt branches, resulting in $O(B^n)$ nodes. If we consider $T$ rounds and do $N$ rollouts to estimate all these nodes, it will be $O(TN B^{nT})$, which is very costly. In practice, we can randomly select one message in one round and randomly sample $M$ nodes from the tree to do the rollout, leading to a cost of $O(TNM)$. 

\subsection{The impact of rollout policies. } As described in Appendix \ref{sec:payoff}, one should interrupt when the current point can reduce the communication cost while maintaining the task performance. In other words, we are optimizing towards an interruption policy that can perform better than the no-interruption baseline, instead of the best interruption behavior. This can mitigate the impact of high variance in rollouts because we only care about the comparison with the no-interruption rather than the exact $Cost$ and $Pref$ value. In our preliminary experiments, we find that estimating with other rollout policies, such as prompting interruption, does not show significant differences with the random policy when getting the comparison result with the no-interruption baseline. Therefore, we select the random rollout policy because it is more computationally efficient and does not rely on any prior hypothesis of the interruption behavior. 

\subsection{Finetuning setting}

We set a hard constraint for the payoff to be positive, which means that each interruption point with label=1 must have a lower communication cost and comparable task performance; there is no hyperparameter tuning of the coefficient to balance cost and performance.  
All our experiments are conducted on the AWS \code{p4de.24xlarge} instances. We train the \method on each task separately and evaluate its performance on this task. 

\subsection{Evaluation setting}
\label{sec:app-evaluation}
We evaluate the communication between Llama 3 70B, 405B, Gemini and the Llama 3 8B, 70B models. For each setting, we run three trials and report the average and standard deviations in Tables \ref{tab:full} and \ref{tab:full_std}. We average the performance over Llama 3 70B, 405B, and Gemini and report the simplified version in \autoref{tab:result}. The temperature is set to the default value of 0.7 during inference.

\begin{LLMPrompt}[Prompting Interruption]
\label{p:cot}
\{ Chat History \} \\
\{ Current Message Chunks\} \\
Given the conversation and the current message, determine if you should interrupt and provide an immediate response. \\
Interrupt if you have enough information to offer a comprehensive reply or if there's a mistake or misunderstanding in the current message. 
\end{LLMPrompt}

\section{More Experimental Results}


\subsection{Probe the agent's understanding level}
\label{sec:app-understanding}
In Section \ref{sec:analysis}, we evaluate LLMs' understanding level to investigate why it interrupts early. We define 5 levels and ask the agent to decide its understanding status based on the current chat history. The prompt used is listed below. 

\begin{LLMPrompt}[Probing Understanding Level]
[Chat History] \\
Now, please provide an estimation on how well you understand the information and the difference. The understanding level can be one of the following: \\
**fully**: Fully understand all details and nuances without needing clarification. \\
**good**: Understand most of the message with minor unclear details. \\
**partial**: Grasp the main topic and some key points, but need clarification on details. \\
**minimal**: Recognize a few words, but the message is mostly unclear. \\
**not at all**: The message is completely unclear. 
Please provide the understanding level in the format of "Understanding: (insert level here)". \\
\end{LLMPrompt}

Meanwhile, we let the agent directly predict the answer given the current context and evaluate the answer's correctness. Then we plot the distribution of the understanding levels for the incorrect answers. We use the GPT-4o as the guesser and use different LLMs as the describer. 
The results are shown in Figure \ref{fig:un_gpt_incorrect}.

\subsection{Hyperparameter Investigation}
\label{sec:app-hyper}
\begin{figure}
    \centering
    \includegraphics[width=1\linewidth]{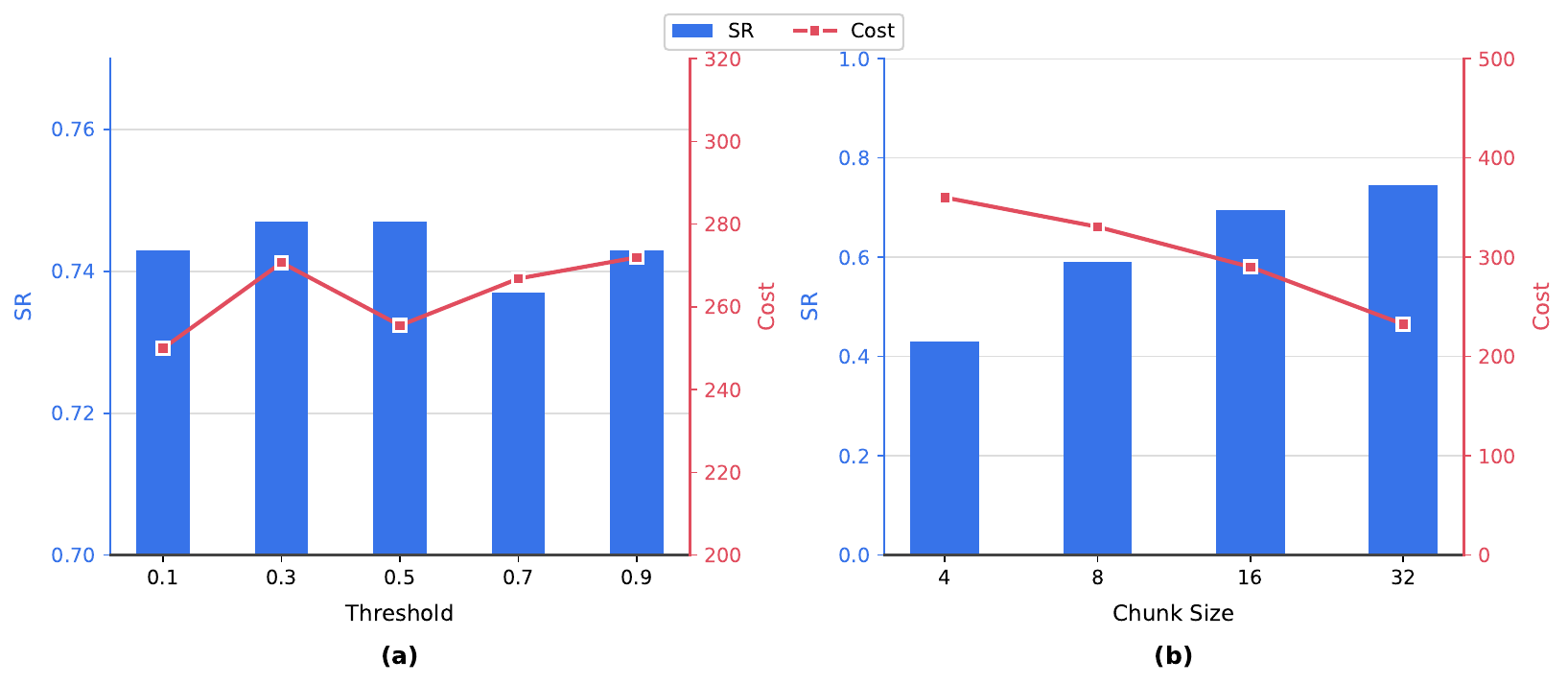}
    \caption{Ablation on different thresholds and chunk sizes. ``SR'' denotes success ratio. }
    \label{fig:thre-chunk}
\end{figure}

\textbf{How sensitive is \method to thresholds?}
Instead of directly using the predicted `Yes' or `No' as the interruption signal, we also investigate the influence of a fine-grained threshold on the token probability. Specifically, we get the probability of the token `Yes' $p_y$ or the token `No' $p_n$. We choose the different thresholds $\theta$ to decide whether to interrupt ($p_y \geq \theta$ or $p_n < \theta$). The results are plotted in \autoref{fig:thre-chunk}. The success rate is similar when the threshold is relaxed, and increases with a stricter threshold. A lower threshold encourages earlier interruption within a round but requires more rounds to complete the task, increasing overall cost. A stricter threshold is more conservative but leads to fewer rounds. This produces a U-shaped cost curve.

\textbf{How will the message chunk size affect the performance?}
A small chunk size requires more interruption checks but may help the \listener interrupt earlier.
We investigate this trade-off by setting different chunk sizes during inference and report the results in \autoref{fig:thre-chunk}. As shown in the figure, a small chunk size does not benefit the success rate or the cost. We find that this is because the agent is not sensitive to small changes in its input: a 4-word increment in the message does not cause much difference in the interruption prediction. This often leads to early interruption. A larger chunk size ensures that key information is not truncated too early and makes it easier for the model to identify the difference between chunks.

\subsection{Common Failure Modes}

We identify three main failure patterns from our experiments based on manual inspection of about 50 incorrect responses.

\begin{enumerate}
\item \textbf{Interrupting too early.} For example, in Text Pictionary, HANDRAISER sometimes interrupts right after the describer says "the previous answer is incorrect," which causes it to miss the subsequent correction that would allow it to guess correctly.

\item \textbf{Unsuitable chunk segmentation.} For example, in Meeting Scheduling, when a planning agent states its available time slots, one chunk may contain only the starting time while the ending time appears in the next chunk. HANDRAISER then interrupts without the ending time and misses a key constraint.

\item \textbf{Interrupting and then asking for continuation.} HANDRAISER sometimes interrupts but then asks the speaker to continue, for example, saying "It seems that your statement is incomplete," rather than directly responding. This negates the cost savings.

\item \textbf{Interrupting too late.} This can also happen, though it is less common than premature interruption. When HANDRAISER is too conservative, the listener waits for most of the message before interrupting, yielding only minor cost savings while task performance remains unchanged. This scenario is more often when the speaker's messages are too short with a large chunk size. We observe that with a smaller chunk size, this scenario is less likely to happen.
\end{enumerate}

\subsection{Combine listener-oriented \method with speaker-oriented compression}

As a first attempt at listener-oriented communication, we demonstrate its promise in reducing communication cost across various LLMs, tasks, and speaker-listener combinations. We do not claim that listener-oriented communication outperforms all speaker-side compression baselines. Rather, the two approaches are orthogonal and can be combined with retraining.

To illustrate this, we conduct additional experiments combining \method with the concise speaker baseline. Results are shown in \autoref{tab:combine}. On average, a compressed speaker further reduces communication cost. However, in some cases, compression hurts task performance, consistent with the Baseline vs. Concise trend in \autoref{tab:result}, suggesting that more advanced speaker compression methods could mitigate this.

\begin{table*}[!htp]\footnotesize\setlength{\tabcolsep}{4pt}
\centering
\caption{Combine listener-oriented \method with speaker-oriented compression. }
\label{tab:combine}
\begin{tabular}{llcccccc}
\toprule
 & &\multicolumn{2}{c}{Textual Pictionary} &\multicolumn{2}{c}{Meeting Scheduling} &\multicolumn{2}{c}{MMLU-Pro-Debate} \\
\cmidrule{3-8}
Listener & Methods & SR & Cost & SR & Cost & SR & Cost \\
\midrule
\multirow{2}[0]{*}{\texttt{Llama-8B}}
    & \method                        & $0.743$ & $262$ & $0.307$ & $1042$ & $0.583$ & $782$ \\
    & \quad + Concise Speaker      & $0.760$ & $253$ & $0.273$ & $1021$ & $0.563$ & $763$ \\
\midrule
\multirow{2}[0]{*}{\texttt{Llama-70B}}
    & \method                        & $0.790$ & $294$ & $0.447$ & $1010$ & $0.657$ & $806$ \\
    & \quad + Concise Speaker      & $0.780$ & $279$ & $0.467$ & $1010$ & $0.653$ & $794$ \\
\bottomrule
\end{tabular}
\end{table*}

\subsection{Full Result Table with Standard Deviations}
We report the full results in \autoref{tab:full}, including separate results for each speaker-listener combination. Standard deviations are in \autoref{tab:full_std}.

\section{Discussion on Duplex Conversational Systems}
\textbf{Duplex Conversational Systems}
Most LLM-based dialogue systems operate in a half-duplex, turn-based fashion, where agents alternate strictly between sending and receiving. To overcome this limitation, duplex models enable simultaneous generation and reception of messages. \citet{zhang2024beyond} propose duplex text LLMs that generate output concurrently with receiving input via a time-division multiplexing strategy, demonstrating improvements in responsiveness and human-likeness. In the speech domain, ~\citep{veluri2024beyond} introduces synchronous LLMs that integrate real-world time information to support full-duplex spoken dialogue with natural turn-taking, overlapping speech, and backchanneling. 

\textbf{Human-Agent Interaction}
In a human-facing setting like tau-bench, one party is a human user whose experience and naturalness of conversation must be preserved~\citep{yao2025taubench,barres2025tau2,qian2025userbench,seshadri2026lost}. These user-facing concerns require additional metrics such as satisfaction scores or naturalness ratings. 

However, our framework is designed for agent-to-agent communication in the backend of multi-agent systems, not for agent-to-user interaction. The key distinction is that the recipient in our setting is an AI agent, not a human. The AI agent does not experience frustration, discomfort, or reduced satisfaction when interrupted. The only relevant cost is whether the communication task is completed successfully and efficiently, which is exactly what our task performance and token cost metrics capture.

Giving the speaker agent the option to reject an interruption signal is a natural extension that would make the framework suitable for human-facing settings; we leave this to future work.

\section{The Use of Large Language Models (LLMs)}
A large language model (LLM) was used as a general-purpose assistive tool to check grammar and correct typographical errors in this paper. The LLM did not contribute to research ideation, experimental design, analysis, or substantive writing. The authors take full responsibility for the content of the paper.

\section{Prompts}
\label{sec:app-task-prompt}
In this section, we list the prompts we use in our three multi-agent scenarios.  

We list the task-specific prompts for each agent in the following. 
\begin{LLMPrompt}[Textual Pictionary]
\textbf{Describer}: You are a describer in a game of text Pictionary. Your task is to describe the word without using the word itself. The word is  answer .\\
\textbf{Guesser}: You are a guesser in a game of text Pictionary. Your task is to guess the word based on the description provided.
\end{LLMPrompt}

\begin{LLMPrompt}[Meeting Scheduling (Planning Agent)]
You are a meeting planner representing a meeting participant.\\ 
You will share information and negotiate with the other planner and the traveler, and finally let the traveler decide.\\ 
\\ 
Each round, you will talk to one of the other agents. You will not talk to yourself. Avoid leaking the detailed personal private information of the meeting participant you represent.\\ 
\\ 
When the meetings the others proposed don't satisfy the constraints or preferences of the meeting participant you represent, you should bravely express your disagreement and articulate the reasons while not leaking the detailed personal private information of your represented meeting participant.\\ 
\\ 
**Personal PUBLIC information**\\ 
You're `planner1` agent who helps to schedule meeting for ** planner1-name **, who is a  planner1-role  in the team.\\ 
 planner1-name  is already at  planner1-location  and will ONLY meet with  traveler-name  here for  planner1-meeting-length  minutes.\\ 
\\ 
 planner1-name  is available at  planner1-available-str . In particular,  planner1-name  prefers to meet at  planner1-preferred-str .\\ 
 planner1-name  already has meetings at  planner1-meetings-str .\\ 
\\ 
**Personal PRIVATE information**\\ 
 planner1-preferences \\ 
\\ 
**Other agents' public information**\\ 
`planner2` agent is helping  planner2-name  to schedule meeting.\\ 
`traveler` agent is helping  traveler-name  to schedule meeting.\\ 
\\ 
**Distance reference**\\ 
 distance-matrix-description \\ 
\\ 
**How to reply**\\ 
Always output a code block containing **exactly two lines**:\\ 
1. **One agent (`planner2` or `traveler`) you will talk to (NOT YOURSELF!)**\\ 
2. **Your full message** (one or more sentences)\\ 
\end{LLMPrompt}

\begin{LLMPrompt}[Meeting Scheduling (Traveler)]
You are a traveler who needs to meet with all the other meeting participants one by one. You will collaborate with the other agents to schedule valid meeting timeline that satisfy the constraints and meeting participant's preferences with the other people.\\ 
\\ 
Each round, you will talk to one of the other agents. You will not talk to yourself.\\ 
\\ 
**Personal information**\\ 
You're `traveler` agent who helps to schedule meeting for ** traveler-name **.\\ 
Today,  traveler-name  will ONLY meet  planner1-name  and  planner2-name  separately.\\ 
\\ 
 traveler-name  will leave  traveler-location  at  starting-time  and drive to the meeting locations.\\ 
\\ 
**Other agents' public information**\\ 
`planner1` agent is helping  planner1-name  to schedule meeting.\\ 
`planner2` agent is helping  planner2-name  to schedule meeting.\\ 
\\ 
**Distance reference**\\ 
 distance-matrix-description \\ 
\\ 
**How to reply**\\ 
Always output a code block containing **exactly two lines**:\\ 
1. **One agent (`planner1` or `planner2`) you will talk to (NOT YOURSELF!)**\\ 
2. **Your full message** (one or more sentences about your thought)\\ 
\\ 
When you are certain all meetings are scheduled:\\ 
1. Respond in a **single fenced code block**.\\ 
2. Inside the block, you should\\ 
• first output **Stop!** as the first line, and\\ 
• then list each scheduled meeting per line in the form:\\ 
    NAME1 - NAME2, START‑TIME – END‑TIME, at LOCATION.
\end{LLMPrompt}

\begin{LLMPrompt}[Multi-agent Debate (Moderator)]
You are a moderator. There will be two debaters involved in a debate. \\ 
They will present their answers and discuss their perspectives on the following topic: [ debate-topic ] \\ 
At the end of each round, you will evaluate answers and decide which is correct. \\ 
You, as the moderator, will evaluate both sides' answers and determine if there is a clear preference for an answer candidate. If so, please summarize your reasons for supporting affirmative/negative side and give the final answer that you think is correct, and the debate will conclude. If not, the debate will continue to the next round. \\ 
For a better evaluation, you can ask the debaters to provide more details and clarify their points. \\ 
Please note that you should not make a decision until you are certain about the correct answer. Therefore, it will be better to wait for at least two interactions between PRO and CON to gather sufficient information.\\ 
Now please output your answer in json format, with the format as follows: ```{"Whether there is a preference": "Yes or No", "Supported Side": "pro or con", "Reason": "", "debate-answer": ""}. ```\\ 
Pay attention, you must include ``` at the beginning and the end of your output.
\\ 
The debate-answer should be either the answer from the pro side or the con side, depending on which side you support. For multi-choice questions, please provide the selected option letter in format of '(X)' without anything else.\\ 
Please strictly output in JSON format, do not output irrelevant content.
\end{LLMPrompt}

\begin{LLMPrompt}[Multi-agent Debate (Debater)]
You are a debater. Hello and welcome to the debate. It's not necessary to fully agree with each other's perspectives, as our objective is to find the correct answer.\\ 
The debate topic is stated as follows:
[ debate-topic ]\\ 
You are the PRO / CON side. \\ 
Your answer is [ answer ]. \\
Your main reason is [ reason ].\\ 
You should first state your stance and then persuade the moderator to support your stance. \\
\end{LLMPrompt}

\begin{table}[htbp]\footnotesize\setlength{\tabcolsep}{1pt}
  \centering
  \caption{Full Results for \autoref{tab:result}. Notations are the same. L8B refers to llama 3 8B, L70B refers to llama 3 70B, and L405B refers to llama 3 405B.}
\resizebox{\textwidth}{!}{%
    \begin{tabular}{clccccc|ccccc}
    \toprule
          &       & \multicolumn{5}{c}{Success Ratio}     & \multicolumn{5}{c}{Communication Cost} \\
    \midrule
          &       & \multicolumn{2}{c}{Noninterruptible} & \multicolumn{3}{c}{Interruptible} & \multicolumn{2}{c}{Noninterruptible} & \multicolumn{3}{c}{Interruptible} \\
    Listener & \multicolumn{1}{c}{Speaker} & \multicolumn{1}{c}{Generic} & \multicolumn{1}{c}{Concise} & \multicolumn{1}{c}{Rand.} & \multicolumn{1}{c}{Prompt} & \multicolumn{1}{c}{\method} & \multicolumn{1}{c}{Generic} & \multicolumn{1}{c}{Concise} & \multicolumn{1}{c}{Rand.} & \multicolumn{1}{c}{Prompt} & \multicolumn{1}{c}{\method} \\
    \midrule
    \multicolumn{12}{c}{Textual Pictionary} \\
    \midrule
    \multirow{4}[1]{*}{L8B} & L70B & 0.670 & 0.710 & 0.690 & 0.400 & 0.710 & 443.8 & 333.8 & 448.8 & 574.3 & 337.0 \\
          & L405B & 0.740 & 0.790 & 0.710 & 0.500 & 0.760 & 361.1 & 290.6 & 386.4 & 513.3 & 280.5 \\
          & Gemini & 0.820 & 0.760 & 0.730 & 0.600 & 0.760 & 232.6 & 206.9 & 245.6 & 296.5 & 168.3 \\
          & \cellcolor[rgb]{ .91,  .91,  .91}AVG & \cellcolor[rgb]{ .91,  .91,  .91}0.743 & \cellcolor[rgb]{ .91,  .91,  .91}0.753 & \cellcolor[rgb]{ .91,  .91,  .91}0.710 & \cellcolor[rgb]{ .91,  .91,  .91}0.500 & \cellcolor[rgb]{ .91,  .91,  .91}0.743 & \cellcolor[rgb]{ .91,  .91,  .91}345.8 & \cellcolor[rgb]{ .91,  .91,  .91}277.1 & \cellcolor[rgb]{ .91,  .91,  .91}360.3 & \cellcolor[rgb]{ .91,  .91,  .91}461.3 & \cellcolor[rgb]{ .91,  .91,  .91}261.9 \\
    \midrule
    \multirow{4}[1]{*}{L70B} & L70B & 0.750 & 0.770 & 0.750 & 0.640 & 0.750 & 462.2 & 395.6 & 507.1 & 557.2 & 285.5 \\
          & L405B & 0.740 & 0.810 & 0.790 & 0.650 & 0.810 & 444.8 & 350.0 & 414.5 & 577.2 & 305.4 \\
          & Gemini & 0.830 & 0.820 & 0.820 & 0.700 & 0.810 & 297.3 & 233.2 & 336.7 & 433.8 & 291.8 \\
          & \cellcolor[rgb]{ .91,  .91,  .91}AVG & \cellcolor[rgb]{ .91,  .91,  .91}0.773 & \cellcolor[rgb]{ .91,  .91,  .91}0.800 & \cellcolor[rgb]{ .91,  .91,  .91}0.787 & \cellcolor[rgb]{ .91,  .91,  .91}0.663 & \cellcolor[rgb]{ .91,  .91,  .91}0.790 & \cellcolor[rgb]{ .91,  .91,  .91}401.4 & \cellcolor[rgb]{ .91,  .91,  .91}326.3 & \cellcolor[rgb]{ .91,  .91,  .91}419.4 & \cellcolor[rgb]{ .91,  .91,  .91}522.7 & \cellcolor[rgb]{ .91,  .91,  .91}294.2 \\
    \midrule
    \multicolumn{12}{c}{Meeting Schedule} \\
    \midrule
    \multirow{4}[1]{*}{L8B} & L70B & 0.290 & 0.250 & 0.240 & 0.240 & 0.340 & 1436.1 & 1281.6 & 1722.2 & 1746.7 & 1167.0 \\
          & L405B & 0.230 & 0.290 & 0.170 & 0.180 & 0.240 & 1267.0 & 1011.8 & 1547.0 & 1647.2 & 946.8 \\
          & Gemini & 0.300 & 0.230 & 0.180 & 0.180 & 0.340 & 1379.2 & 1172.9 & 2009.6 & 1939.4 & 1012.4 \\
          & \cellcolor[rgb]{ .91,  .91,  .91}AVG & \cellcolor[rgb]{ .91,  .91,  .91}0.273 & \cellcolor[rgb]{ .91,  .91,  .91}0.257 & \cellcolor[rgb]{ .91,  .91,  .91}0.197 & \cellcolor[rgb]{ .91,  .91,  .91}0.200 & \cellcolor[rgb]{ .91,  .91,  .91}0.307 & \cellcolor[rgb]{ .91,  .91,  .91}1360.7 & \cellcolor[rgb]{ .91,  .91,  .91}1155.4 & \cellcolor[rgb]{ .91,  .91,  .91}1759.6 & \cellcolor[rgb]{ .91,  .91,  .91}1777.8 & \cellcolor[rgb]{ .91,  .91,  .91}1042.0 \\
    \midrule
    \multirow{4}[1]{*}{L70B} & L70B & 0.370 & 0.490 & 0.450 & 0.450 & 0.490 & 1287.9 & 1082.3 & 1446.1 & 1813.2 & 1068.3 \\
          & L405B & 0.440 & 0.370 & 0.390 & 0.440 & 0.390 & 1219.5 & 1071.6 & 1749.3 & 1812.4 & 1045.6 \\
          & Gemini & 0.450 & 0.450 & 0.410 & 0.370 & 0.460 & 1175.8 & 920.6 & 1560.5 & 1730.8 & 918.6 \\
          & \cellcolor[rgb]{ .91,  .91,  .91}AVG & \cellcolor[rgb]{ .91,  .91,  .91}0.420 & \cellcolor[rgb]{ .91,  .91,  .91}0.437 & \cellcolor[rgb]{ .91,  .91,  .91}0.417 & \cellcolor[rgb]{ .91,  .91,  .91}0.420 & \cellcolor[rgb]{ .91,  .91,  .91}0.447 & \cellcolor[rgb]{ .91,  .91,  .91}1227.7 & \cellcolor[rgb]{ .91,  .91,  .91}1024.8 & \cellcolor[rgb]{ .91,  .91,  .91}1585.3 & \cellcolor[rgb]{ .91,  .91,  .91}1785.5 & \cellcolor[rgb]{ .91,  .91,  .91}1010.8 \\
    \midrule
    \multicolumn{12}{c}{MMLU-Pro-Debate} \\
    \midrule
    \multirow{4}[1]{*}{L8B} & L70B & 0.510 & 0.530 & 0.500 & 0.550 & 0.570 & 1746.2 & 835.4 & 1946.2 & 2083.3 & 802.4 \\
          & L405B & 0.550 & 0.580 & 0.560 & 0.550 & 0.620 & 1462.4 & 858.6 & 1526.8 & 1697.5 & 813.6 \\
          & Gemini & 0.500 & 0.460 & 0.550 & 0.510 & 0.560 & 1531.2 & 794.5 & 1690.3 & 1964.7 & 732.5 \\
          & \cellcolor[rgb]{ .91,  .91,  .91}AVG & \cellcolor[rgb]{ .91,  .91,  .91}0.520 & \cellcolor[rgb]{ .91,  .91,  .91}0.523 & \cellcolor[rgb]{ .91,  .91,  .91}0.537 & \cellcolor[rgb]{ .91,  .91,  .91}0.537 & \cellcolor[rgb]{ .91,  .91,  .91}0.583 & \cellcolor[rgb]{ .91,  .91,  .91}1580.0 & \cellcolor[rgb]{ .91,  .91,  .91}829.5 & \cellcolor[rgb]{ .91,  .91,  .91}1721.1 & \cellcolor[rgb]{ .91,  .91,  .91}1915.2 & \cellcolor[rgb]{ .91,  .91,  .91}782.8 \\
    \midrule
    \multirow{4}[1]{*}{L70B} & L70B & 0.660 & 0.680 & 0.650 & 0.670 & 0.660 & 1831.1 & 1113.3 & 1803.1 & 2302.6 & 1042.3 \\
          & L405B & 0.700 & 0.650 & 0.660 & 0.690 & 0.680 & 1552.1 & 817.5 & 1650.2 & 2050.0 & 786.5 \\
          & Gemini & 0.580 & 0.640 & 0.560 & 0.600 & 0.630 & 1466.4 & 609.4 & 1668.1 & 1619.0 & 589.4 \\
          & \cellcolor[rgb]{ .91,  .91,  .91}AVG & \cellcolor[rgb]{ .91,  .91,  .91}0.647 & \cellcolor[rgb]{ .91,  .91,  .91}0.657 & \cellcolor[rgb]{ .91,  .91,  .91}0.623 & \cellcolor[rgb]{ .91,  .91,  .91}0.653 & \cellcolor[rgb]{ .91,  .91,  .91}0.657 & \cellcolor[rgb]{ .91,  .91,  .91}1616.5 & \cellcolor[rgb]{ .91,  .91,  .91}846.7 & \cellcolor[rgb]{ .91,  .91,  .91}1707.1 & \cellcolor[rgb]{ .91,  .91,  .91}1990.6 & \cellcolor[rgb]{ .91,  .91,  .91}806.1 \\
    \bottomrule
    \end{tabular}%
}
  \label{tab:full}%
\end{table}%

\begin{table}[htbp]\footnotesize\setlength{\tabcolsep}{2pt}
  \centering
  \caption{STD for \autoref{tab:full}. Notations are the same. L8B refers to llama 3 8B, L70B refers to llama 3 70B, and L405B refers to llama 3 405B.}
\resizebox{\textwidth}{!}{%
    \begin{tabular}{clccccc|ccccc}
        \toprule
          &       & \multicolumn{5}{c}{Success Ratio}     & \multicolumn{5}{c}{Communication Cost} \\
    \midrule
          &       & \multicolumn{2}{c}{Noninterruptible} & \multicolumn{3}{c}{Interruptible} & \multicolumn{2}{c}{Noninterruptible} & \multicolumn{3}{c}{Interruptible} \\
    Listener & \multicolumn{1}{c}{Speaker} & \multicolumn{1}{c}{Generic} & \multicolumn{1}{c}{Concise} & \multicolumn{1}{c}{Rand.} & \multicolumn{1}{c}{Prompt} & \multicolumn{1}{c}{\method} & \multicolumn{1}{c}{Generic} & \multicolumn{1}{c}{Concise} & \multicolumn{1}{c}{Rand.} & \multicolumn{1}{c}{Prompt} & \multicolumn{1}{c}{\method} \\
    \midrule
    \multirow{4}[1]{*}{L8B} & L70B & 0.06  & 0.01  & 0.03  & 0.03  & 0.018 & 42.9  & 29.0  & 37.0  & 18.8  & 27.8 \\
          & L405B & 0     & 0.05  & 0.05  & 0.02  & 0.024 & 40.2  & 28.1  & 17.5  & 32.7  & 32.4 \\
          & Gemini-2.0 & 0.04  & 0.01  & 0.02  & 0.05  & 0.028 & 28.8  & 16.3  & 12.6  & 23.6  & 19.5 \\
          & \cellcolor[rgb]{ .91,  .91,  .91}AVG & \cellcolor[rgb]{ .91,  .91,  .91}0.033 & \cellcolor[rgb]{ .91,  .91,  .91}0.023 & \cellcolor[rgb]{ .91,  .91,  .91}0.033 & \cellcolor[rgb]{ .91,  .91,  .91}0.033 & \cellcolor[rgb]{ .91,  .91,  .91}0.023 & \cellcolor[rgb]{ .91,  .91,  .91}37.3 & \cellcolor[rgb]{ .91,  .91,  .91}24.5 & \cellcolor[rgb]{ .91,  .91,  .91}22.4 & \cellcolor[rgb]{ .91,  .91,  .91}25.0 & \cellcolor[rgb]{ .91,  .91,  .91}26.6 \\
    \multirow{4}[1]{*}{L70B} & L70B & 0.03  & 0.03  & 0.03  & 0.02  & 0.03  & 61.6  & 38.3  & 33.7  & 52.9  & 46.9 \\
          & L405B & 0.03  & 0.01  & 0.04  & 0.04  & 0.02  & 65.0  & 42.9  & 30.4  & 45.2  & 59.4 \\
          & Gemini-2.0 & 0.01  & 0.04  & 0.03  & 0.06  & 0.01  & 12.7  & 30.2  & 28.7  & 31.5  & 40.5 \\
          & \cellcolor[rgb]{ .91,  .91,  .91}AVG & \cellcolor[rgb]{ .91,  .91,  .91}0.023 & \cellcolor[rgb]{ .91,  .91,  .91}0.027 & \cellcolor[rgb]{ .91,  .91,  .91}0.033 & \cellcolor[rgb]{ .91,  .91,  .91}0.040 & \cellcolor[rgb]{ .91,  .91,  .91}0.020 & \cellcolor[rgb]{ .91,  .91,  .91}46.4 & \cellcolor[rgb]{ .91,  .91,  .91}37.2 & \cellcolor[rgb]{ .91,  .91,  .91}30.9 & \cellcolor[rgb]{ .91,  .91,  .91}43.2 & \cellcolor[rgb]{ .91,  .91,  .91}48.9 \\
    \midrule
    \multicolumn{12}{c}{Meeting Schedule} \\
    \midrule
    \multirow{4}[1]{*}{L8B} & L70B & 0.08  & 0.02  & 0.05  & 0.01  & 0.048 & 100.5 & 69.1  & 155.9 & 139.2 & 84.8 \\
          & L405B & 0.07  & 0.06  & 0.03  & 0.01  & 0.052 & 49.4  & 48.1  & 37.2  & 100.8 & 88.3 \\
          & Gemini-2.0 & 0.08  & 0.08  & 0.04  & 0.04  & 0.048 & 109.0 & 121.9 & 149.7 & 148.2 & 74.6 \\
          & \cellcolor[rgb]{ .91,  .91,  .91}AVG & \cellcolor[rgb]{ .91,  .91,  .91}0.077 & \cellcolor[rgb]{ .91,  .91,  .91}0.053 & \cellcolor[rgb]{ .91,  .91,  .91}0.040 & \cellcolor[rgb]{ .91,  .91,  .91}0.020 & \cellcolor[rgb]{ .91,  .91,  .91}0.049 & \cellcolor[rgb]{ .91,  .91,  .91}86.3 & \cellcolor[rgb]{ .91,  .91,  .91}79.7 & \cellcolor[rgb]{ .91,  .91,  .91}114.3 & \cellcolor[rgb]{ .91,  .91,  .91}129.4 & \cellcolor[rgb]{ .91,  .91,  .91}82.6 \\
    \multirow{4}[1]{*}{L70B} & L70B & 0.030 & 0.04  & 0.080 & 0.030 & 0.100 & 53.5  & 70.2  & 55.9  & 55.0  & 57.2 \\
          & L405B & 0.040 & 0.06  & 0.060 & 0.090 & 0.040 & 44.8  & 122.0 & 158.0 & 174.5 & 114.9 \\
          & Gemini-2.0 & 0.050 & 0.03  & 0.100 & 0.060 & 0.050 & 56.8  & 54.0  & 89.5  & 180.0 & 90.0 \\
          & \cellcolor[rgb]{ .91,  .91,  .91}AVG & \cellcolor[rgb]{ .91,  .91,  .91}0.040 & \cellcolor[rgb]{ .91,  .91,  .91}0.043 & \cellcolor[rgb]{ .91,  .91,  .91}0.080 & \cellcolor[rgb]{ .91,  .91,  .91}0.060 & \cellcolor[rgb]{ .91,  .91,  .91}0.063 & \cellcolor[rgb]{ .91,  .91,  .91}51.7 & \cellcolor[rgb]{ .91,  .91,  .91}82.1 & \cellcolor[rgb]{ .91,  .91,  .91}101.1 & \cellcolor[rgb]{ .91,  .91,  .91}136.5 & \cellcolor[rgb]{ .91,  .91,  .91}87.4 \\
    \midrule
    \multicolumn{12}{c}{MMLU-Pro-Debate} \\
    \midrule
    \multirow{4}[1]{*}{L8B} & L70B & 0.020 & 0.05  & 0.030 & 0.060 & 0.086 & 202.9 & 314.3 & 106.7 & 52.1  & 156.0 \\
          & L405B & 0.090 & 0.09  & 0.020 & 0.060 & 0.074 & 86.9  & 136.9 & 29.7  & 160.5 & 159.3 \\
          & Gemini-2.0 & 0.040 & 0.13  & 0.100 & 0.070 & 0.080 & 86.7  & 77.7  & 118.0 & 176.5 & 228.6 \\
          & \cellcolor[rgb]{ .91,  .91,  .91}AVG & \cellcolor[rgb]{ .91,  .91,  .91}0.050 & \cellcolor[rgb]{ .91,  .91,  .91}0.090 & \cellcolor[rgb]{ .91,  .91,  .91}0.050 & \cellcolor[rgb]{ .91,  .91,  .91}0.063 & \cellcolor[rgb]{ .91,  .91,  .91}0.080 & \cellcolor[rgb]{ .91,  .91,  .91}125.5 & \cellcolor[rgb]{ .91,  .91,  .91}176.3 & \cellcolor[rgb]{ .91,  .91,  .91}84.8 & \cellcolor[rgb]{ .91,  .91,  .91}129.7 & \cellcolor[rgb]{ .91,  .91,  .91}181.3 \\
    \multirow{4}[1]{*}{L70B} & L70B & 0.020 & 0.030 & 0.000 & 0.030 & 0.100 & 59.5  & 35.6  & 215.2 & 114.0 & 114.6 \\
          & L405B & 0.010 & 0.080 & 0.040 & 0.080 & 0.030 & 39.0  & 102.3 & 110.4 & 201.7 & 105.7 \\
          & Gemini-2.0 & 0.010 & 0.060 & 0.090 & 0.110 & 0.080 & 134.4 & 199.9 & 130.3 & 253.2 & 137.4 \\
          & \cellcolor[rgb]{ .91,  .91,  .91}AVG & \cellcolor[rgb]{ .91,  .91,  .91}0.013 & \cellcolor[rgb]{ .91,  .91,  .91}0.057 & \cellcolor[rgb]{ .91,  .91,  .91}0.043 & \cellcolor[rgb]{ .91,  .91,  .91}0.073 & \cellcolor[rgb]{ .91,  .91,  .91}0.070 & \cellcolor[rgb]{ .91,  .91,  .91}77.6 & \cellcolor[rgb]{ .91,  .91,  .91}112.6 & \cellcolor[rgb]{ .91,  .91,  .91}152.0 & \cellcolor[rgb]{ .91,  .91,  .91}189.6 & \cellcolor[rgb]{ .91,  .91,  .91}119.3 \\
    \bottomrule
    \end{tabular}%
}
  \label{tab:full_std}%
\end{table}%

\end{document}